\documentclass[12pt]{iopart}

\usepackage{iopams}
\usepackage{bm,color}  
\usepackage{graphicx}
\usepackage{amssymb}

\newcommand{\T}[1]{\tilde{#1}}

\newcommand{\be}{\begin{eqnarray}}
\newcommand{\ee}{\end{eqnarray}}
\newcommand{\ba}{\begin{array}}
\newcommand{\ea}{\end{array}}

\newcommand{\subbe}{\begin{subequations}}
\newcommand{\subee}{\end{subequations}}

\begin{document}

\title[Perfect reconstruction of sparse signals with nonconvex penalties]
{Perfect reconstruction of sparse signals with 
piecewise continuous nonconvex penalties and nonconvexity control}
\author{Ayaka Sakata$^{1,2,3}$ and Tomoyuki Obuchi$^{4}$}

\address{$^1$Department of Statistical Inference \& Mathematics, The Institute of Statistical Mathematics, Midori-cho, Tachikawa, Tokyo 190-8562, Japan}
\address{$^2$Department of Statistical Science, The Graduate University for Advanced Science (SOKENDAI), Hayama-cho, Kanagawa 240-0193, Japan}
\address{$^3$JST PRESTO,
4-1-8 Honcho, Kawaguchi, Saitama 332-0012, Japan}
\address{$^4$Graduate School of Informatics, Kyoto University,
36-1 Yoshida-Honmachi, Sakyo-ku, Kyoto 606-8501, Japan}
\ead{$^{1,2,3}$ayaka@ism.ac.jp,$^{4}$obuchi@i.kyoto-u.ac.jp}
\address{}
\vspace{10pt}
\begin{abstract}
We consider 
a reconstruction problem of sparse signals
from smaller number of measurements than the dimension
formulated as a minimization problem of nonconvex sparse penalties, 
Smoothly Clipped Absolute Deviation (SCAD)
and Minimax Concave Penalty (MCP). The nonconvexity of these penalties is controlled by 
nonconvexity parameters, and $\ell_1$ penalty is contained as a limit
with respect to these parameters.
The analytically derived reconstruction limit overcomes
that of $\ell_1$ 
limit, and is also expected to does the
algorithmic limit of the Bayes-optimal setting,
when the nonconvexity parameters have suitable values.
However, for small nonconvexity parameters,
where the reconstruction of the relatively dense signals is 
theoretically expected, the algorithm known as
approximate message passing (AMP),
which is closely related to the analysis,
cannot 
achieve perfect reconstruction
leading to a gap from the analysis.
Using the theory of state evolution, it is clarified that this gap can be understood 
on the basis of the shrinkage in the basin of attraction to 
the perfect reconstruction and also the divergent behavior of AMP in some regions.
A part of the gap is mitigated by
controlling the shapes of nonconvex penalties
to guide the AMP trajectory to the basin of the attraction.
\end{abstract}

\section{Introduction}

A signal processing scheme for reconstructing signals through linear measurements,
when the number of measurements is less than the dimensionality of the signals, is
known as compressed sensing (or compressive sensing) \cite{Candes-Tao,Donoho2006}.
Let $\bm{x}^0\in\mathbb{R}^N$ and $A\in\mathbb{R}^{M\times N}$ 
be the unknown original signal and 
measurement matrix, respectively.
Compressed sensing is mathematically formulated as a problem of reconstructing the
signal $\bm{x}^0$ from its measurement
$\bm{y}=A\bm{x}^0$, where
the number of measurements is less than the signal dimension ($M<N$).
In general, the problem is underdetermined, and the solution is not unique. However, 
the signal can be reconstructed utilizing the knowledge that 
the original signal is sparse; it contains zero components
with a finite probability.
The reconstruction of signals from a limited number of measurements
is a common challenge in various fields.
In the past decade, theories and techniques of compressed sensing 
have been enriched by the interdisciplinary work in the fields 
such as signal processing, medical imaging, and statistical physics.

A natural way to reconstruct a sparse signal is 
to minimize the $\ell_0$ norm under a constraint:
\begin{equation}
\min_{\bm{x}}||\bm{x}||_0,~\mathrm{subject~to~}\bm{y}=A\bm{x},
\label{eq:CS_L0}
\end{equation}
where $||\bm{x}||_0$ is the number of nonzero components in $\bm{x}$.
However, a combinatorial search with
respect to the support set is required to exactly solve \eref{eq:CS_L0};
hence, it is unrealistic for implementation.
The minimization of $\ell_1$ 
norm \cite{Candes-Tao,basis_pursuit} is a widely used approach:
\begin{equation}
\min_{\bm{x}}||\bm{x}||_1,~\mathrm{subject~to~}\bm{y}=A\bm{x},
\label{eq:CS_L1}
\end{equation}
which is a convex relaxation problem of \eref{eq:CS_L0},
where $||\bm{x}||_1=\sum_{i=1}^N|x_i|$.
Efficient algorithms to solve \eref{eq:CS_L1} have been developed \cite{OMP,DMM},
in addition to the convex optimization techniques \cite{IRLS}.
Further, some conditions about the measurement matrix, such as the null space property and the restricted isometry property \cite{NP,RIP}, are found and under such conditions it is shown that the solutions \eref{eq:CS_L0} and \eref{eq:CS_L1} become equivalent. 

Despite the mathematical tractability of $\ell_1$ minimization,
its performance is inferior to $\ell_0$ minimization 
for the practical setting of the measurement matrix $A$.
The difference between $\ell_1$ and $\ell_0$ is expected to be 
reduced by introducing the 
minimization of $\ell_p~(0<p< 1)$ norm.
In fact, $\ell_p~(0<p<1)$ minimization
achieves the reconstruction of the original signal
from a fewer number of measurements
than $\ell_1$ minimization \cite{Lp_CS,Lp_RIP}.
However, $\ell_p~(0<p<1)$ minimization leads to a discontinuity of the reconstructed
signal with respect to the input,
which induces algorithmic instability.
Smoothly Clipped Absolute Deviation (SCAD) \cite{SCAD} and 
Minimax Concave Penalty (MCP) \cite{MCP},
which are piecewise continuous nonconvex penalties, are potential candidates to address this limitation.
SCAD and MCP are designed to 
provide continuity, unbiasedness, and sparsity to the estimates,
and their nonconvexities are controlled by nonconvexity parameters.
The mathematical treatment of nonconvex penalties is seemingly difficult compared with $\ell_1$.
However, it is shown that
a data compression problem under SCAD and MCP
can be solved without additional computational cost 
compared with convex optimization problems 
in a certain parameter region,
and this region is characterized by replica symmetry
in the context of statistical physics \cite{Sakata-Xu}.
This investigation implies 
prospects of these penalties for improvement in 
reconstructing the signals in 
compressed sensing.

In this study, we theoretically verify the performance of the 
minimization of SCAD and MCP
for the reconstruction of sparse signals
in compressed sensing.
The perfect reconstruction is achieved 
with a smaller number of measurements
compared with $\ell_1$ reconstruction limit \cite{Donoho_typical,Donoho-Tanner,CS_replica}.
Further, SCAD and MCP minimization overcomes
the algorithmic limit of the Bayes-optimal method \cite{Krzakala2012}, 
in the sense that there exists 
a unique stable solution corresponding to the perfect reconstruction 
even beyond the algorithmic limit of the Bayes-optimal method, 
and that there is no phase transitions which can be algorithmic barriers. 
Based on this finding, as a reconstruction algorithm, 
we employ the approximated message passing (AMP) algorithm
to SCAD and MCP minimization.
Examining its performance, it is found that 
AMP cannot actually achieve the perfect reconstruction beyond the Bayes-optimal algorithmic limit,
despite the theoretical basis.
To investigate the gap
between the AMP's behavior and the analytical result, 
we use the technique of the state evolution (SE), which allows us to track 
the macroscopic dynamics of AMP.
As a result, 
it is found that the gap comes from the shrinkage 
in the basin of attraction to the perfect reconstruction
and the divergent behavior of AMP in some parameter regions. 
One of the contribution of the paper is the finding of the 
scenario of the algorithmic failure different from that in the Bayes-optimal setting
where an emergence of another local minimum 
than the success solution, which is the global minimum, hampers
the convergence of AMP to 
the perfect reconstruction.

We mitigate the abovementioned gap and 
improve the performance of AMP by introducing a method 
controlling nonconvexity parameters, 
named nonconvexity control, into AMP, which is our another contribution. 
The efficiency of the nonconvexity control is 
understood from the flow of SE. Further, 
the property of the fixed point of SE gives a guide for the protocol 
of the nonconvexity control. The algorithmic limit of SCAD and MCP minimization, 
called nonconvexity control limit, 
is determined as the limit where the proposed 
nonconvexity control leads to the perfect reconstruction. 
The resultant algorithmic limit is very close to 
but slightly inferior to the Bayes-optimal algorithmic limit.

The remainder of this paper is organized as follows.
In Sec. \ref{sec:def_SCAD_MCP},
we introduce the nonconvex sparse penalties, SCAD and MCP, used in this study.
The equilibrium properties of compressed sensing
under SCAD and MCP are studied in Sec. \ref{sec:replica}, 
based on the replica method under replica symmetric (RS) assumption.
In Sec. \ref{sec:success},
the limit for the perfect reconstruction is derived for SCAD and MCP,
and we show that their performance 
is expected to overcome 
$\ell_1$ reconstruction limit and the algorithmic limit of the Bayes-optimal method
by investigating the presence of phase transitions and the local stability of the 
solution corresponding to the perfect reconstruction.
In Sec. \ref{sec:AMP}, we demonstrate the actual reconstruction of the signal 
using AMP, and show that
the reconstructed signal diverges when the nonconvexity parameters are small,
even when the reconstruction is theoretically supported.
We show that this divergence can be suppressed by introducing 
the nonconvexity control.
Sec. \ref{sec:summary} is devoted to the summary and discussion of this paper.

\section{Definition of SCAD and MCP}
\label{sec:def_SCAD_MCP}

The problem considered in this study is formulated as 
\begin{equation}
\min_{\bm{x}}J(\bm{x};\lambda,a)~\mbox{subject to}~\bm{y}=A\bm{x},
\label{eq:CS_def}
\end{equation}
where
$J(\bm{x};\lambda,a)=\sum_{i=1}^NJ(x_i;\lambda,a)$ is a sparsity-inducing penalty
and $\lambda, a$ are regularization parameters.
We deal with two types of nonconvex penalties, SCAD and MCP.
The shapes of these penalties are controlled by two parameters $\lambda$ and $a$,
and $\ell_1$ penalty is considered as a limit.
We call these regularization parameters 
nonconvexity parameters.

\begin{figure}
\centering
\includegraphics[width=3in]{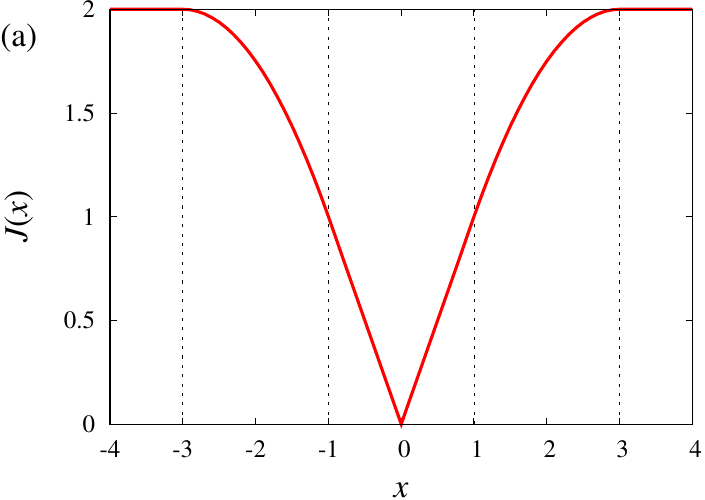}
\includegraphics[width=3in]{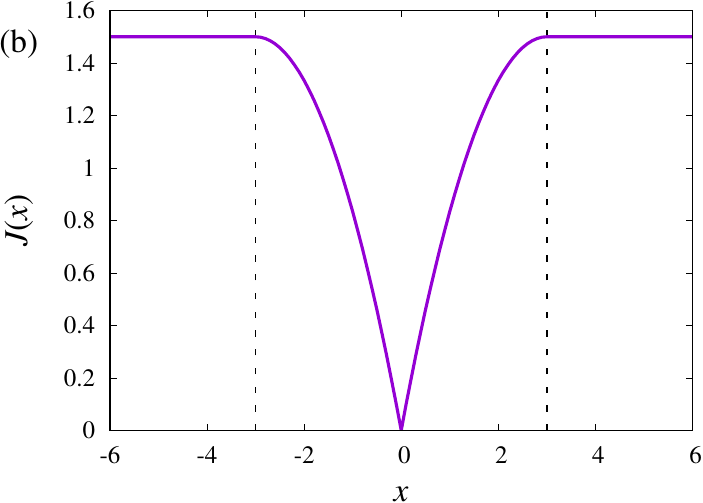}
\caption{Shapes of (a) SCAD for $\lambda=1$ and $a=3$ and (b) MCP for $\lambda=1$ and $a=3$.
The dashed lines represent the thresholds where penalty shape changes.}
\label{fig:SCAD_and_MCP}
\end{figure}

SCAD is defined by \cite{SCAD}
\begin{equation}
J(x;\lambda,a)=\left\{\begin{array}{ll}
\lambda|x| & (|x|\leq \lambda) \\
-\displaystyle\frac{x^2-2a\lambda|x|+\lambda^2}{2(a-1)} & (\lambda<|x|\leq a\lambda) \\
\displaystyle\frac{(a+1)\lambda^2}{2} & (|x|>a\lambda)
\end{array}
\right.,
\label{eq:def_SCAD}
\end{equation}
where $\lambda\in(0,\infty)$ and $a\in(1,\infty)$.
\Fref{fig:SCAD_and_MCP} (a) represents SCAD penalty 
at $\lambda=1$ and $a=3$.
The dashed vertical lines are the thresholds 
$|x|=\lambda$ and $|x|=a\lambda$.
SCAD penalty for $|x|\leq\lambda$ is equivalent to $\ell_1$ penalty,
and that for $|x|>a\lambda$ is equivalent to 
$\ell_0$ penalty; i.e., the penalty has a constant value.
These $\ell_1$ and $\ell_0$ regions are 
connected with each other through
a quadratic function.
At $a\to\infty$, SCAD is reduced to $\ell_1$ penalty, $J(x;\lambda,a\to\infty)=\lambda|x|$,
and the minimization of SCAD at $\lambda\to\infty$ is also equivalent to 
that of $\ell_1$ minimization.

MCP is defined by \cite{MCP}
\begin{eqnarray}
J(x;\lambda,a)&=\left\{\begin{array}{ll}
\lambda|x|-\displaystyle\frac{x^2}{2a} & (|x|\leq a\lambda) \\
\displaystyle\frac{a\lambda^2}{2} & (|x|> a\lambda) 
\end{array}
\right.,
\end{eqnarray}
where $\lambda\in(0,\infty)$ and $a\in(1,\infty)$.
\Fref{fig:SCAD_and_MCP} (b) represents MCP for $\lambda=1$ and $a=3$.
The vertical line represents the threshold $|x|=a\lambda$.
As with SCAD,
MCP is also reduced to $\ell_1$ by taking the limit $a\to\infty$,
{and also equivalent to $\ell_1$ minimization when
the penalty is minimized at $\lambda\to\infty$.}

\subsection{Estimator under SCAD and MCP}
\label{sec:SCAD_MCP_def}

\begin{figure}
\begin{minipage}{0.333\hsize}
\centering
\includegraphics[width=1.8in]{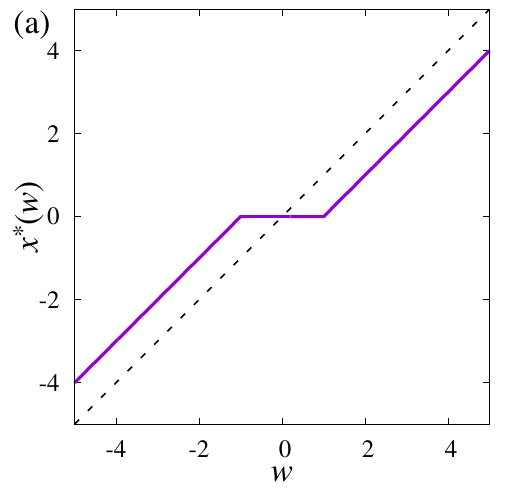}
\end{minipage}
\begin{minipage}{0.333\hsize}
\centering
\includegraphics[width=1.8in]{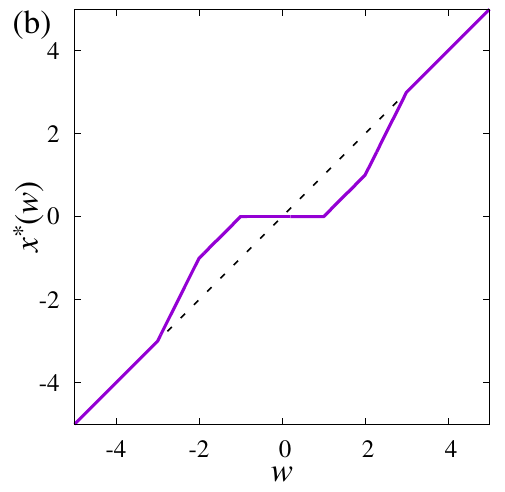}
\end{minipage}
\begin{minipage}{0.333\hsize}
\centering
\includegraphics[width=1.8in]{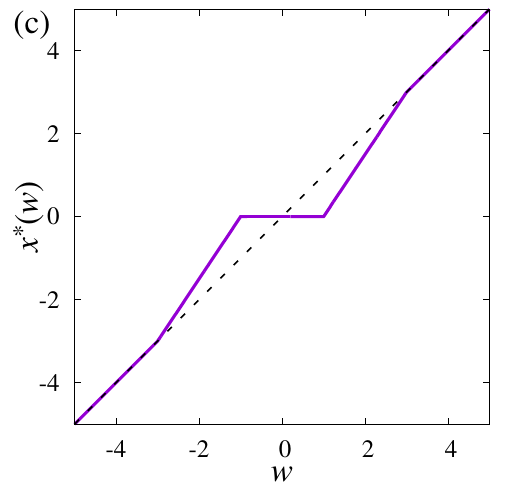}
\end{minipage}
\caption{Behaviour of the estimator for (a) $\ell_1$, (b) SCAD, and (c) MCP.
The dashed diagonal lines represent OLS estimator.}
\label{fig:estimators}
\end{figure}

For understanding the properties of SCAD and MCP,
let us consider the one-dimensional fitting problem 
of the input $w$ using Gaussian model
penalized by SCAD or MCP as
\begin{eqnarray}
\hat{x}(s,w)=\arg\min_x\left\{\frac{x^2}{2s}-wx+J(x;\lambda,a)\right\},
\label{eq:one-dim-ex}
\end{eqnarray}
where $s>0$.
Both of SCAD and MCP have upward convex terms, hence
to define a solution of \eref{eq:one-dim-ex} for all $w$ region,
we need to carefully consider the possible region of $s$ and $a$,
which give the coefficient of the quadratic term in R.H.S. of \eref{eq:one-dim-ex}.
In case of SCAD, 
the relationship $a>1+s$ should hold to
obtain the solution as 
\begin{eqnarray}
\hat{x}(s,w)=\Sigma_{\mathrm{SCAD}}(s,w){\cal M}_{\mathrm{SCAD}}(s,w)
\label{eq:one-dim-SCAD}
\end{eqnarray}
where ${\cal M}_{\mathrm{SCAD}}$ and $\Sigma_{\mathrm{SCAD}}$ represent 
the coefficients of linear and quadratic term in \eref{eq:one-dim-ex} given by
\begin{eqnarray}
{\cal M}_{\mathrm{SCAD}}(s,w)&=\left\{\begin{array}{ll}
w-\mathrm{sgn}(w)\lambda & \mathrm{for}~\lambda(1+s^{-1})\geq|w|>\lambda\\
w-\mathrm{sgn}(w)\frac{a\lambda}{a-1} & \mathrm{for}~a\lambda s^{-1}\geq|w|>\lambda(1+s^{-1})\\
w & \mathrm{for}~|w|>a\lambda s^{-1}\\
0 & \mathrm{otherwise}
\end{array}
\right.,\\
\Sigma_{\mathrm{SCAD}}(s,w)&=\left\{\begin{array}{ll}
s & \mathrm{for}~\lambda(1+s^{-1})\geq|w|>\lambda\\
\left(s^{-1}-\frac{1}{a-1}\right)^{-1}& \mathrm{for}~a\lambda s^{-1}\geq|w|>\lambda(1+s^{-1})\\
s & \mathrm{for}~|w|>a\lambda s^{-1}\\
0 & \mathrm{otherwise}
\end{array}
\right.,
\end{eqnarray}
and $\mathrm{sgn}(w)$ denotes the sign of $w$.
An example of the estimator as a function of the input $w$
under SCAD is shown in \Fref{fig:estimators} (b),
where $s=1$, $\lambda=1$ and $a=3$.
The SCAD estimator behaves like the $\ell_1$ estimator,
which is shown in \Fref{fig:estimators} (a), and like the ordinary least square (OLS)
estimator when 
$\lambda(1+s^{-1})\geq|z|>\lambda$ and $|w|>a\lambda s^{-1}$, respectively. In the region $a\lambda s^{-1}\geq|w|>\lambda(1+s^{-1})$,
the estimator linearly transits between $\ell_1$ and OLS estimators.

From the same argument as SCAD,
$a>s$ should hold to define the solution of \eref{eq:one-dim-ex}
under MCP. Further, the condition $a>1$ is imposed in the definition of MCP,
hence in summarize the restriction is given as $a>\min\{1,s\}$.
When the condition $a>\min\{1,s\}$ is satisfied,
the solution of the single body problem under MCP is given by \cite{Sakata2018}
\begin{equation}
x^*(s,w)=\Sigma_{\mathrm{MCP}}(s,w){\cal M}_{\mathrm{MCP}}(s,w),
\end{equation}
where
\begin{eqnarray}
{\cal M}_{\mathrm{MCP}}(s,w)&=\left\{\begin{array}{ll}
w-\mathrm{sgn}(w)\lambda & \mathrm{for}~a\lambda s^{-1}\geq|w|>\lambda\\
w & \mathrm{for}~|w|>a\lambda s^{-1}\\
0 & \mathrm{otherwise}
\end{array}
\right.,\\
\Sigma_{\mathrm{MCP}}(s,w)&=\left\{\begin{array}{ll}
(s^{-1}-a^{-1})^{-1} & \mathrm{for}~a\lambda s^{-1}\geq|w|>\lambda\\
s & \mathrm{for}~|w|>a\lambda s^{-1}\\
0 & \mathrm{otherwise}
\end{array}
\right..
\label{eq:MCP_one_body}
\end{eqnarray}
\Fref{fig:estimators} (c) shows the behaviour of the MCP estimator at $s=1$, $\lambda=1$, and $a=3$.
The MCP estimator behaves like the OLS estimator at $|w|>a\lambda s^{-1}$,
and is connected from zero to the OLS estimator in the region $a\lambda s^{-1}\geq|w|>\lambda$. 

\section{Replica analysis for SCAD and MCP}
\label{sec:replica}

We assume that the signal to be reconstructed is generated according to the 
Bernoulli--Gaussian distribution,
\begin{equation}
P_0(\bm{x}^0)=\prod_i\left\{(1-\rho)\delta(x_i^0)+\frac{\rho}{\sqrt{2\pi\sigma_x^2}}\exp\left(-\frac{(x_i^0)^2}{2\sigma_x^2}\right)\right\},
\end{equation}
where $\delta(x)$ is the Dirac delta function.
Further, we consider the measurement matrix $A$ to be a random Gaussian,
where each component is independently and identically distributed 
according to the Gaussian distribution with
mean 0 and variance $N^{-1}$. 
The measurement is expressed as $\bm{y}=A\bm{x}^0$,
and the minimization of $J(\bm{x};\lambda,a)$ is 
implemented under the constraint $\bm{y}=A\bm{x}$.
For mathematical tractability,
we express the constraint $\bm{y}=A\bm{x}$ by introducing a parameter $\tau$ as
\begin{equation}
P_\tau(\bm{y}|\bm{x})=\frac{1}{(\sqrt{2\pi\tau})^M}\exp\left\{-\frac{1}{2\tau}||\bm{y}-A\bm{x}||_2^2\right\},
\end{equation}
where the probability is concentrated at $\bm{y}=A\bm{x}$ 
taking the limit $\tau\to 0$.
The posterior distribution corresponding to the problem \eref{eq:CS_def}
is given by
\begin{equation}
P(\bm{x}|\bm{y})=\lim_{\beta\to\infty}\lim_{\tau\to 0}\frac{1}{Z_{\beta,\tau}(\bm{y})}\exp(-\beta J(\bm{x};\lambda,a))P_\tau(\bm{y}|\bm{x}),
\label{eq:posterior}
\end{equation}
where $\beta$ is a parameter to attain
the uniform distribution over the minimizer of \eref{eq:CS_def} at $\beta\to\infty$,
and
\begin{equation}
    Z_{\beta,\tau}(\bm{y})=\int d\bm{x}\exp(-\beta J(\bm{x};\lambda,a))P_\tau(\bm{y}|\bm{x})
\end{equation}
is the normalization constant.
The minimizer of \eref{eq:CS_def} is given by
$\hat{\bm{x}}=\langle\bm{x}\rangle$,
where $\langle\cdot\rangle$ denotes the expectation with respect to $\bm{x}$ according to the 
posterior distribution \eref{eq:posterior}.

The performance of the reconstruction
\eref{eq:CS_def} depends on the randomness $A$ and $\bm{x}^0$. Here,
we examine the typical performance of the SCAD and MCP minimization
at $N\to\infty$ and $M\to\infty$
keeping $M\slash N=\alpha\sim O(1)$,
where $\alpha$ is the compression ratio.
Free energy density defined by
\begin{equation}
f=-\lim_{\beta\to\infty}\lim_{N\to\infty}\lim_{\tau\to 0}\frac{1}{N\beta}E_{\bm{x^0},A}[\ln Z_{\beta,\tau}(\bm{y})]
\label{eq:f_def}
\end{equation}
is the key in this discussion,
where $E_{\bm{x}^0,A}[\cdots]$ denotes the expectation with respect to $A$ and $\bm{x}^0$
introduced for the discussion of the typical property.
Here, we proceed the calculation for general $\tau$ and $\beta$, and 
take the limit after the derivation of the general form.
It is calculated using
the following identity
\begin{equation}
E_{\bm{x}^0,A}[\ln Z_{\beta,\tau}(\bm{y})]=\lim_{n\to 0}\frac{E_{\bm{x}^0,A}[Z^n_{\beta,\tau}(\bm{y})]-1}{n}.
\label{eq:replica}
\end{equation}
Assuming that $n$ is a positive integer,
we can express the expectation of $Z_\beta^n$
in \eref{eq:replica}
using $n$-replicated systems
\begin{eqnarray}
\nonumber
E_{\bm{x}^0,A}[Z_{\beta,\tau}^n(\bm{y})]&=\int dAd\bm{y}d\bm{x}^0P_0(\bm{x}^0)P_{A}(A)\delta(\bm{y}-A\bm{x}^0)\\
\nonumber
&\times\int 
 d\bm{x}^{(1)}\cdots d\bm{x}^{(n)}\lim_{\tau\to 0}\frac{1}{(\sqrt{2\pi\tau})^{nM}}\\
&\times\exp\Big[\sum_{a=1}^n\Big\{-\frac{1}{2\tau}||\bm{y}-A\bm{x}^{(a)}||_2^2-\beta J(\bm{x}^{(a)};\lambda,a)\Big\}\Big].
\label{eq:E_replica}
\end{eqnarray}
The detail of the calculation is shown in \cite{Sakata2016, Sakata-Xu,CS_replica},
and here we briefly summarize the calculation.
The free energy density 
under the RS assumption is given by
\begin{equation}
f=\mathrm{extr}_{\Omega,\T{\Omega}}\left[\frac{\alpha(Q-2m+\rho\sigma_x^2)}{2\chi}+m\T{m}-\frac{\T{Q}Q-\T{\chi}\chi}{2}+\frac{\overline{\xi(\T{Q},\sigma)}}{2}\right],
\end{equation}
where $\mathrm{extr}_{\Omega,\T{\Omega}}$ represents the extremization with 
respect to the quantities
$\Omega=\{Q,\chi,m\}$ and $\tilde{\Omega}=\{\tilde{Q},\T{\chi},\tilde{m}\}$.
The function $\xi(\tilde{Q},\sigma)$ depends on the regularization as
\begin{eqnarray}
\xi(\T{Q},\sigma)&=2\int Dz L(\T{Q},\sigma z),\label{eq:f_xi_def}
\\
L(\T{Q},\sigma z)&=\min_x\Big(\frac{\T{Q}}{2}x^2-\sigma zx+J(x;\lambda,a)\Big),\label{eq:one_body}
\end{eqnarray}
where $\int Dz=\int_{-\infty}^{\infty}dz\exp(-z^2\slash 2)\slash\sqrt{2\pi}$.
\Eref{eq:one_body}
is equivalent to the one-dimensional problem \eref{eq:one-dim-ex}.
Here, 
$\overline{\cdots}$ denotes the average over 
$\sigma$ according to the distribution
\begin{equation}
P_\sigma(\sigma)=(1-\rho)\delta(\sigma-\sigma_-)+\rho\delta(\sigma-\sigma_+),
\label{eq:sigma_dist}
\end{equation}
with $\sigma_-=\sqrt{\T{\chi}}$ and $\sigma_+=\sqrt{\T{\chi}+\T{m}^2\sigma_x^2}$.
The random fields $\sigma_- z$ and $\sigma_+z$ effectively represents the
randomness induced by $A$ and $\bm{x}^0$,
in particular zero-signals and non-zero-signals, respectively.
We denote the solution of $x$ in the effective single-body problem \eref{eq:one_body}
as $x^*(\T{Q}^{-1},\sigma z)$, 
which depends on the regularization,
and we consider the specific form of $x^*(\T{Q}^{-1},\sigma z)$ 
and $L(\tilde{Q},\sigma z)$ later.
The saddle point equations are given by
\begin{eqnarray}
\chi&=-\frac{\partial\overline{\xi(\T{Q},\sigma)}}{\partial\T{\chi}}=\int Dz \overline{\frac{\partial x^*(\T{Q}^{-1},\sigma z)}{\partial(\sigma z)}},\label{eq:RS_chi_gen}\\
Q&=\frac{\partial\overline{\xi(\T{Q},\sigma)}}{\partial\T{Q}}
=\int Dz \overline{(x^*(\T{Q}^{-1},\sigma z))^2},\label{eq:RS_Q_gen}\\
m&=-\frac{1}{2}\frac{\partial\overline{\xi(\T{Q},\sigma)}}{\partial\T{m}}
={\rho\T{m}\sigma_x^2}\int Dz\frac{\partial x^*(\T{Q}^{-1},\sigma_+z)}{\partial (\sigma_+z)},
\label{eq:RS_m_gen}\\
\T{\chi}&=\frac{\alpha(Q-2m+\rho\sigma_x^2)}{\chi^2},\label{eq:RS_chih}\\
\T{Q}&=\frac{\alpha}{\chi},\label{eq:RS_Qh}\\
\T{m}&=\frac{\alpha}{\chi}.\label{eq:RS_mh}
\end{eqnarray}
At the saddle point, $x^*(\T{Q}^{-1},\sigma z)$ is statistically equivalent to the
point estimate $\hat{\bm{x}}$, and 
$\chi,Q$ and $m$ are related to the physical quantities as 
\begin{eqnarray}
Q&=\lim_{N\to\infty}\frac{1}{N}\sum_{i=1}^NE_{\bm{x}^0,A}[\hat{x}_i^2],\label{eq:Q_physical}\\
m&=\lim_{N\to\infty}\frac{1}{N}\sum_{i=1}^NE_{\bm{x}^0,A}[x_i^0 \hat{x}_i],\label{eq:m_physical}\\
\chi&=\lim_{\beta\to\infty}\lim_{N\to\infty} \frac{\beta}{N}\sum_{i=1}^NE_{\bm{x}^0,A}\left[\langle x_i^2\rangle-\langle x_i\rangle^2\right].\label{eq:chi_physical}
\end{eqnarray}
Hence, the expectation value of 
the mean squared error (MSE) between the reconstructed signal and the original signal
is represented as
\begin{equation}
\varepsilon\equiv\frac{1}{N}E_{\bm{x}^0,A}\left[||\hat{\bm{x}}-\bm{x}^0||_2^2\right]=Q-2m+\rho\sigma_x^2.
\label{eq:MSE}
\end{equation}
The saddle point equations for variables $\Omega=\{Q,\chi,m\}$ directly
depend on the functional form of the regularization,
but the equations for $\T{\Omega}$ do not depend on it.
In the following subsections,
we show the form of saddle point equations of $\Omega$ for SCAD and MCP.

The RS solution 
is stable against the symmetry breaking perturbation
when \cite{Sakata-Xu,CS_replica}
\begin{equation}
\frac{\alpha}{\chi^2}\int Dz\overline{\left(\frac{\partial x^*(\T{Q}^{-1},\sigma z)}{\partial (\sigma z)}\right)^2}<1,
\label{eq:AT}
\end{equation}
which is known as de Almeida--Thouless (AT) condition \cite{AT}.
The form of this condition also depends on the functional form of the
regularization.

\subsection{SCAD}

As mentioned in Sec.\ref{sec:SCAD_MCP_def}, 
$\T{Q}>(a-1)^{-1}$ should hold to define the minimizer of \eref{eq:one_body}.
In the following, the subspace of the 
macroscopic parameters
where the minimizer of \eref{eq:one_body} can be
defined is denoted by $\Omega^\dag(a)\equiv\{Q,\chi,m|\tilde{Q}>(a-1)^{-1}\}$ for each $a$,
and we restrict our discussion to $\Omega^\dag(a)$.
When we are in $\Omega^\dag(a)$,
the minimizer of \eref{eq:one_body}
under SCAD is given by \cite{Sakata2018}
\begin{equation}
x^*(\T{Q}^{-1},\sigma z)=\Sigma_{\mathrm{SCAD}}(\T{Q}^{-1},\sigma z){\cal M}_{\mathrm{SCAD}}(\T{Q}^{-1},\sigma z),
\label{eq:x_SCAD}
\end{equation}
and substituting the solution \eref{eq:x_SCAD} into  
\eref{eq:one_body}, we obtain
\begin{equation}
-2L(\T{Q},\sigma z)=\left\{\begin{array}{ll}
\displaystyle\frac{(\sigma z-\lambda{\rm sgn}(z))^2}{\T{Q}} &  (\sqrt{2}\theta_1(\sigma)<|z|\leq \sqrt{2}\theta_2(\sigma))\\
\displaystyle\frac{(\sigma z-\frac{a\lambda}{a-1})^2}{\T{Q}-\frac{1}{a-1}}+\frac{\lambda^2}{a-1} &
 (\sqrt{2}\theta_2(\sigma)<|z|\leq \sqrt{2}\theta_3(\sigma))\\
\displaystyle\frac{(\sigma z)^2}{\T{Q}}-(a+1)\lambda^2 & (|z|>\sqrt{2}\theta_3(\sigma))\\
0& ({\rm otherwise})
\end{array}
\right.,
\label{eq:single_body_SCAD}
\end{equation}
where $\theta_1(\sigma)={\lambda}\slash(\sqrt{2}\sigma)$, $\theta_2(\sigma)=\lambda(1+\T{Q})\slash(\sqrt{2}\sigma)$, and $\theta_3(\sigma)=a\lambda\T{Q}\slash(\sqrt{2}\sigma)$.
\Eref{eq:f_xi_def} for SCAD regularization is derived as 
\begin{eqnarray}
-\xi(\sigma)=\xi_1(\sigma)+\xi_2(\sigma)+\xi_3(\sigma)+\frac{\lambda^2\xi_4(\sigma)}{a-1}-(a+1)\lambda
 ^2\mathrm{erfc}(\theta_{3}(\sigma)),
\end{eqnarray}
where
\begin{eqnarray}
\nonumber
\xi_1(\sigma)&=\frac{\sigma^2}{\T{Q}}\Big[-\frac{2\theta_{1}(\sigma)}{\sqrt{\pi}}\Big(e^{-\theta_{1}^2(\sigma)}+(\T{Q}-1)e^{-\theta_{2}^2(\sigma)}\Big)\\
&+(1+2\theta_{1}^2(\sigma))\{\mathrm{erfc}(\theta_1(\sigma))-{\rm erfc}(\theta_{2}(\sigma))\}\Big],\\
\nonumber
\xi_2(\sigma)&=\frac{\sigma^2}{\T{Q}-\frac{1}{a-1}}\Big[\frac{2}{\sqrt{\pi}}\Big\{\theta_{2}(\sigma)e^{-\theta_{2}^2(\sigma)}\\
\nonumber
&-\theta_{3}(\sigma)e^{-\theta_{3}^2(\sigma)}-\frac{2\theta_{3}(\sigma)}{\T{Q}(a-1)}\left(e^{-\theta_{2}^2(\sigma)}-e^{-\theta_{3}^2(\sigma)}\right)\Big\}\\
&+\Big\{1+2\Big(\frac{\theta_{3}(\sigma)}{\T{Q}(a-1)}\Big)^2\Big\}\xi_4(\sigma)\Big],\\
\xi_3(\sigma)&=\frac{\sigma^2}{\T{Q}}\Big[\frac{2\theta_{3}(\sigma)}{\sqrt{\pi}}e^{-\theta_{3}^2(\sigma)}+{\rm erfc}(\theta_{3}(\sigma))\Big],\\
\xi_4(\sigma)&={\rm erfc}(\theta_{2}(\sigma))-{\rm erfc}(\theta_{3}(\sigma)).
\end{eqnarray}
The regularization-dependent saddle point equations are given by
\begin{eqnarray}
Q&=\overline{\frac{\xi_1(\sigma)}{\T{Q}}+\frac{\xi_2(\sigma)}{\T{Q}-\frac{1}{a-1}}+\frac{\xi_3(\sigma)}{\T{Q}}},\label{eq:Q_SCAD}\\
\chi&=\frac{1}{\T{Q}}\Big[\hat{\rho}+\frac{\frac{1}{a-1}}{\T{Q}-\frac{1}{a-1}}\overline{\xi_4(\sigma)}\Big],
\label{eq:chi_SCAD}\\
m&={\rho\sigma_x^2}\left[\mathrm{erfc}(\theta_1(\sigma_+))+\frac{\frac{1}{a-1}\xi_4(\sigma_+)}{\T{Q}-\frac{1}{a-1}}\right],
\end{eqnarray}
where $\hat{\rho}$ is the density of the nonzero component in the estimate given by
\begin{equation}
\hat{\rho}=\overline{\mathrm{erfc}(\theta_1(\sigma))}.
\end{equation}
From \eref{eq:AT}, the AT condition is derived as
\begin{equation}
\frac{1}{\alpha}\left[\hat{\rho}+\left\{\left(\frac{\T{Q}}{\T{Q}\!-\!\frac{1}{a-1}}\right)^2-1\right\}\overline{\xi_4(\sigma)}\right]<1.
\label{eq:AT_SCAD}
\end{equation}

\subsection{MCP}

As with SCAD, we concentrate our discussion on
the 
subspace of the macroscopic parameters $\Omega^\dag(a)=\{Q,\chi,m|\tilde{Q}>a^{-1}\}$,
where the solution of \eref{eq:one_body} can be defined.
The solution of the single body problem under MCP 
in $\Omega^\dag(a)=\{Q,\chi,m|\tilde{Q}>a^{-1}\}$
is given by \cite{Sakata2018}
\begin{equation}
x^*(\T{Q}^{-1},\sigma z)=\Sigma_{\mathrm{MCP}}(\T{Q}^{-1},\sigma z){\cal M}_{\mathrm{MCP}}(\T{Q}^{-1},\sigma z),
\end{equation}
and we obtain
\begin{equation}
-2L(\T{Q},\sigma)=\left\{\begin{array}{ll}
\displaystyle\frac{(\sigma z-{\lambda{\rm sgn}(z))^2}}{\T{Q}-a^{-1}} &
 (\sqrt{2}\theta_1(\sigma)<|z|\leq \sqrt{2}\theta_2(\sigma))\\
\displaystyle\frac{(\sigma z)^2}{\T{Q}}-\lambda a^2 & (|z|>\sqrt{2}\theta_2(\sigma))\\
0& ({\rm otherwise})
\end{array}
\right.,
\label{eq:single_body_MCP}
\end{equation}
where $\theta_{1}(\sigma)=\lambda\slash(\sqrt{2}\sigma)$ and
$\theta_{2}(\sigma)=a\lambda\T{Q}\slash(\sqrt{2}\sigma)$ , and
\eref{eq:f_xi_def} for MCP is derived as
\begin{eqnarray}
-\xi(\sigma) = \xi_1(\sigma)+\xi_2(\sigma),
\end{eqnarray}
where
\begin{eqnarray}
\nonumber
\xi_1(\sigma)&=-\frac{2\sigma^2}{\sqrt{\pi}(\T{Q}-a^{-1})}
\left\{\theta_1(\sigma) (e^{-\theta_1^2(\sigma)}-e^{-\theta_2^2(\sigma)})-e^{-\theta_2^2(\sigma)}(\theta_1(\sigma)-\theta_2(\sigma))\right\}\\
&+\frac{(\sigma^2+\lambda^2)\xi_3(\sigma)}{\T{Q}-a^{-1}},\\
\xi_2(\sigma)&=\frac{\sigma^2}{\T{Q}}\left(\frac{2\theta_2(\sigma)}{\sqrt{\pi}}e^{-\theta_2^2(\sigma)}
+{\rm erfc}(\theta_2(\sigma))\right)-\lambda a^2{\rm erfc}(\theta_2(\sigma)),\\
\xi_3(\sigma)&={\rm erfc}(\theta_1(\sigma))-{\rm erfc}(\theta_2(\sigma)).
\end{eqnarray}
The saddle point equations for $\Omega$ are given by
\begin{eqnarray}
Q&=\overline{\frac{\xi_1(\sigma)}{\T{Q}-\frac{1}{a}}
+\frac{\sigma^2}{\T{Q}^2}\left\{\frac{2\theta_2(\sigma)}{\sqrt{\pi}}
e^{-\theta_2^2(\sigma)}+{\rm erfc}(\theta_2(\sigma))\right\}},\\
\chi&=\frac{1}{\T{Q}}\left[\hat{\rho}+\frac{a^{-1}\overline
{\xi_3(\sigma)}}{\T{Q}-a^{-1}}\right],\label{eq:chi_MCP}\\
m&=\rho\sigma_x^2\left[\mathrm{erfc}(\theta_1(\sigma_+))+\frac{a^{-1}\xi_3(\sigma_+)}{\T{Q}-a^{-1}}\right],
\end{eqnarray}
where $\hat{\rho}$ is the density of nonzero component in the estimate given by
\begin{eqnarray}
\hat{\rho}=\overline{{\rm erfc}(\theta_1(\sigma))}.
\end{eqnarray}
The AT condition for MCP is derived as
\begin{eqnarray}
\nonumber
\frac{1}{\alpha}\left[\hat{\rho}+\left\{\left(\frac{\T{Q}}{\T{Q}-a^{-1}}\right)^2-1\right\}\overline{\xi_3(\sigma)}\right]<1.\\
\end{eqnarray}

\section{Stability of success solution}
\label{sec:success}

One of the solutions of the saddle point equations 
in $\Omega^\dag(a)$
is characterized by $Q=m=\rho\sigma_x^2$.
Following the correspondence between the order parameters and the MSE
\eref{eq:MSE},
this solution indicates the perfect reconstruction of the original signal $\bm{x}^0$.
Hence, we call the solution with $Q=m=\rho\sigma_x^2$ the {\it success solution}.
The saddle point equation can have solutions other than the success solution;
however, these solutions do not satisfy the AT condition 
as far as we observed.
Substituting the relationship $Q=m=\rho\sigma_x^2$,
we immediately obtain $\chi=0$ and $\T{Q}=\T{m}=\infty$,
and the only variable to be solved is $\T{\chi}$.
The expansion of $Q$ and $m$ up to the order $O(\T{Q}^{-2})$
gives the expression of $\T{\chi}$ for the success solution
under SCAD
\begin{eqnarray}
\nonumber
\T{\chi}&=\frac{1-\rho}{\alpha}\left[-\frac{2\T{\chi}}{\sqrt{\pi}}\theta_-e^{-\theta_-^2}+(\T{\chi}+\lambda^2)\mathrm{erfc}(\theta_-)\right]\\
\nonumber
&+\frac{\rho}{\alpha}
\Big[\T{\chi}+\lambda^2\left\{1-\mathrm{erfc}(\theta_+)\right\}+\left\{\left(\frac{a\lambda}{a-1}\right)^2+\frac{\sigma_x^2}{(a-1)^2}\right\}\{\mathrm{erfc}(\theta_+)-\mathrm{erfc}(a\theta_+)\}\\
&+\frac{2\sigma_x^2\theta_+}{\sqrt{\pi}(a-1)}\left\{\frac{a}{a-1}(e^{-a^2\theta_+^2}-e^{-\theta_+^2})-e^{-\theta_+^2}\right\}\Big],
\label{eq:chih_success_SCAD}
\end{eqnarray}
and under MCP
\begin{eqnarray}
\nonumber
\T{\chi}&=\frac{1-\rho}{\alpha}\left\{-\frac{2\T{\chi}}{\sqrt{\pi}}\theta_- e^{-\theta_-^2}+(\T{\chi}+\lambda^2)\mathrm{erfc}(\theta_-)\right\}\\
&+\frac{\rho}{\alpha}\left[\T{\chi}+\left(\lambda^2+\frac{\sigma_x^2}{a^2}\right)\left(1-\mathrm{erfc}(a\theta_+)\right)+\frac{2\sigma_x^2\theta_+}{a\sqrt{\pi}}e^{-a^2\theta_+^2}-\frac{4\sigma_x^2\theta_+}{a\sqrt{\pi}}\right],
\label{eq:chih_success_MCP}
\end{eqnarray}
where $\theta_-=\lambda\slash\sqrt{2\T{\chi}}$ and
$\theta_+=\lambda\slash\sqrt{2\sigma_x^2}$.
Eqs. (\ref{eq:chih_success_SCAD}) and (\ref{eq:chih_success_MCP}) are reduced to 
the saddle point equation of $\T{\chi}$ corresponding to the success solution for $\ell_1$  
regularization by setting $\lambda=1$ and when $a\to\infty$ \cite{CS_replica}.

For both the penalties,
the success solution is a locally stable solution
as a saddle point of the RS free energy 
when 
\begin{equation}
\frac{1}{\alpha}\left\{(1-\rho)\mathrm{erfc}\left(\theta_-\right)+\rho\right\}<1.
\label{eq:AT_success}
\end{equation} 
This condition is derived by the linear stability analysis of $\chi$ around 0.
Further, we can show that the AT condition for the success solution is equivalent to \eref{eq:AT_success}.
This means that when the success solution is locally stable as a RS saddle point,
it is also stable with respect to the replica symmetry breaking perturbation.
Therefore, the reconstruction limit $\alpha_c(\rho)$ is defined as
the minimum value of $\alpha$ that satisfies \eref{eq:AT_success}
for each $\rho$.
We also define $\rho_c(\alpha)$ as 
the maximum value of $\rho$ to satisfy \eref{eq:AT_success}
at $\alpha$, and we use both $\alpha_c(\rho)$ and $\rho_c(\alpha)$ for convenience.

\begin{figure}
\begin{minipage}{0.495\hsize}
\centering
\includegraphics[width=3in]{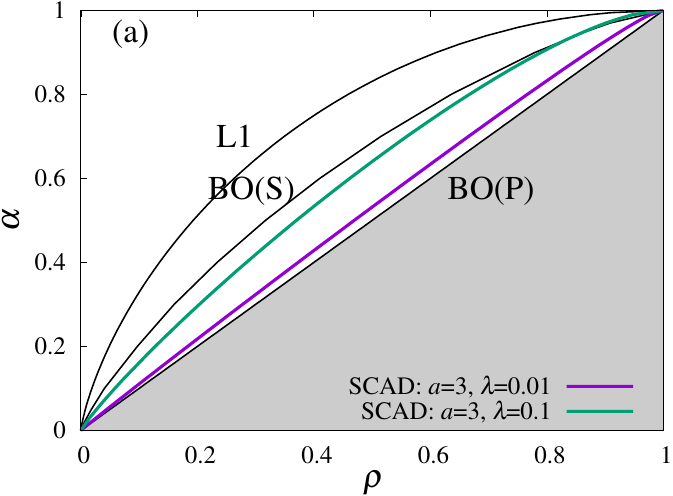}
\end{minipage}
\begin{minipage}{0.495\hsize}
\includegraphics[width=3in]{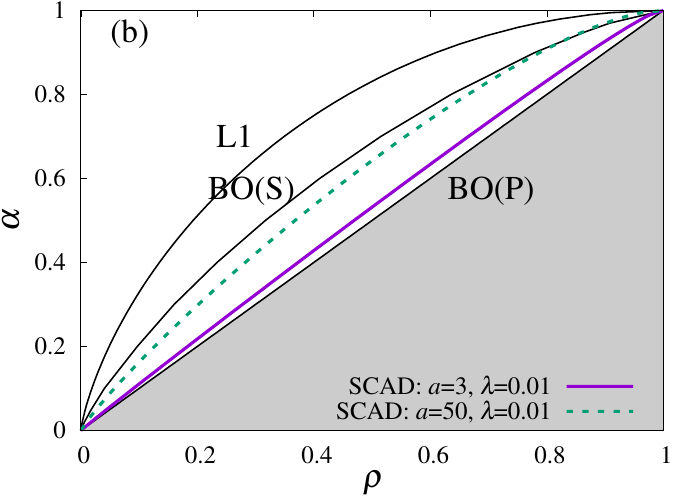}
\end{minipage}
\caption{Reconstruction limit of SCAD at (a) $a=3$ for $\lambda=0.01$ and $\lambda=0.1$ and (b) $\lambda=0.01$ $a=3$ and $a=50$. 
The lines with `L1',
`BO(S)' and `BO(P)' are the reconstruction limit
under $\ell_1$ minimization, the algorithmic limit by the 
Bayes-optimal method given by spinodal transition, 
and the phase transition point of the Bayes-optimal method, respectively.
The shaded regions are $\alpha<\rho$.}
\label{fig:SCAD_phase}
\end{figure}

\begin{figure}
\begin{minipage}{0.495\hsize}
\centering
\includegraphics[width=3in]{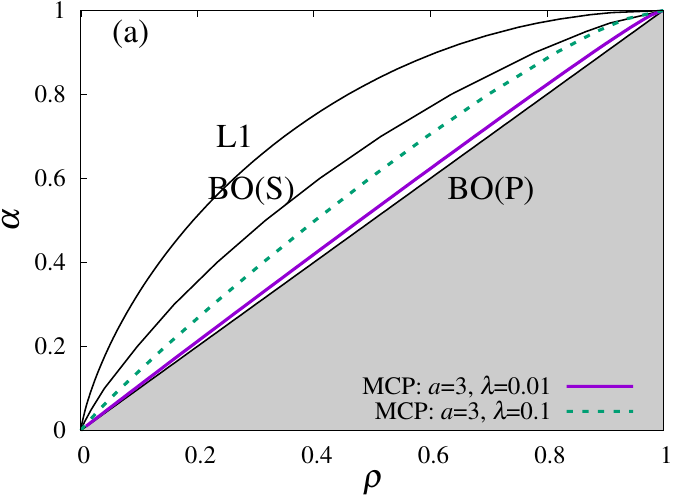}
\end{minipage}
\begin{minipage}{0.495\hsize}
\includegraphics[width=3in]{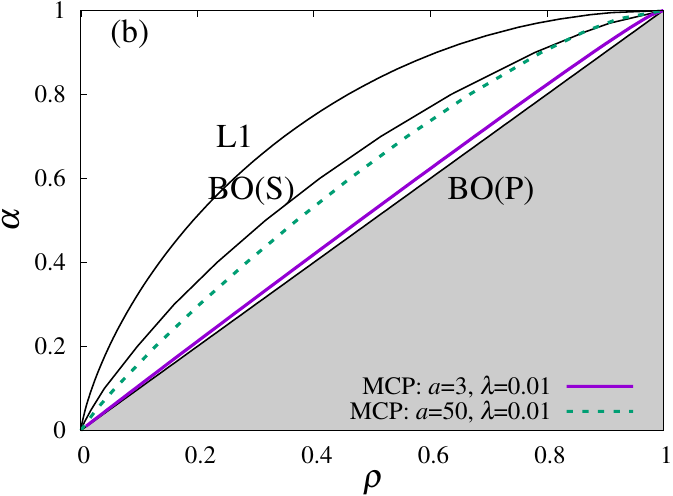}
\end{minipage}
\caption{Reconstruction limit of MCP at (a) $a=3$ for $\lambda=0.01$ and $\lambda=0.1$
and (b) $\lambda=0.01$ for $a=3$ and $a=50$. 
The lines with `L1',
`BO(S)' and `BO(P)' are the same as \Fref{fig:SCAD_phase}.
The shaded regions are $\alpha<\rho$.}
\label{fig:MCP_phase}
\end{figure}

Figs. \ref{fig:SCAD_phase} and \ref{fig:MCP_phase} show 
the $\rho$-dependence of $\alpha_c(\rho)$ for SCAD minimization
and MCP minimization, respectively,
where (a)s represent $a=3$ 
and (b)s represent $\lambda=0.01$.
The typical reconstruction is possible in the parameter region
$\alpha\geq\alpha_c(\rho)$,
and that for $\ell_1$ minimization (L1) and 
the algorithmic limit of the Bayes-optimal method
given by spinodal transition (BO(S)),
and the phase transition boundary of the Bayes-optimal method (BO(P))
over which the success solution is locally stable, are
shown for comparison.
As $\lambda$ and $a$ decrease, $\alpha_c(\rho)$ of SCAD and MCP
become less than that of the algorithmic limit of 
the Bayes-optimal reconstruction method \cite{Krzakala2012}.
Further, 
the reconstruction limit $\alpha_c(\rho)$ approaches $\rho$ as $\lambda\to0$.
Mathematically, $\alpha_c(\rho)\to\rho$ is provided by scaling 
$\theta_-\to\infty$ and $\T{\chi}\to 0$ at $\lambda\to 0$,
which reduces \eref{eq:AT_success} to $\rho<\alpha$.
This inequality, $\rho < \alpha$, is considered to be the 
fundamental
limit, because in        
general sparse estimation methods, 
the estimation of the support increases the effective degrees of the estimated variables.
Hence, we need more measurements than the number of the variables to be estimated.
It is indicated that SCAD and MCP with $\lambda\to 0$
achieve the typical reconstruction
when the number of the measurements and 
that of nonzero variables are balanced.

\begin{figure}
\begin{minipage}{0.495\hsize}
\centering
\includegraphics[width=3in]{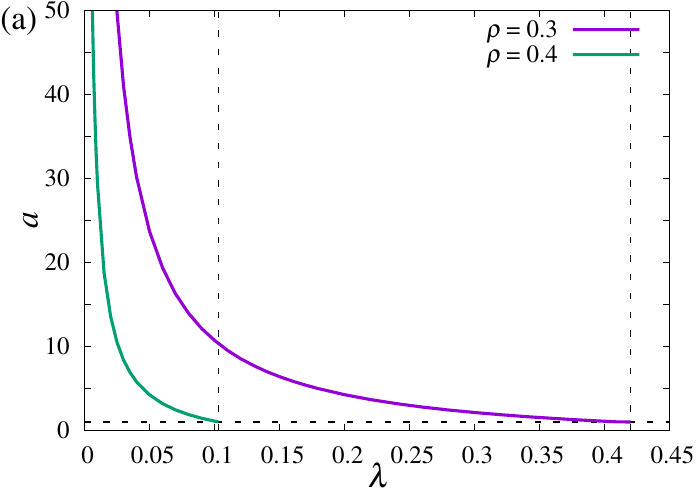}
\end{minipage}
\begin{minipage}{0.495\hsize}
\centering
\includegraphics[width=3in]{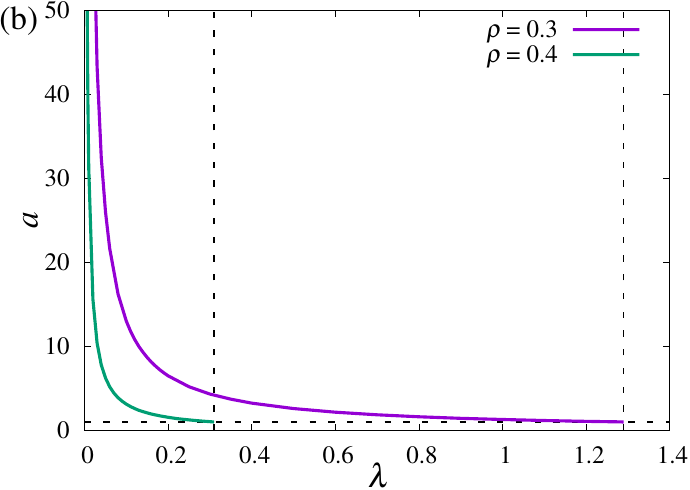}
\end{minipage}
\caption{Reconstruction limit $a_c(\lambda)$ at $\alpha=0.5$
for (a) SCAD and (b) MCP, respectively.
The vertical dashed lines represent the maximum value of $\lambda$,
where the reconstruction is possible with $a_{\min}<a\leq a_c(\lambda)$.
The horizontal dashed lines represent $a=1$,
which is the minimum value of $a$ for the success solution.}
\label{fig:phase_lambda_vs_a}
\end{figure}

\begin{figure}
\begin{minipage}{0.495\hsize}
\centering
\includegraphics[width=3in]{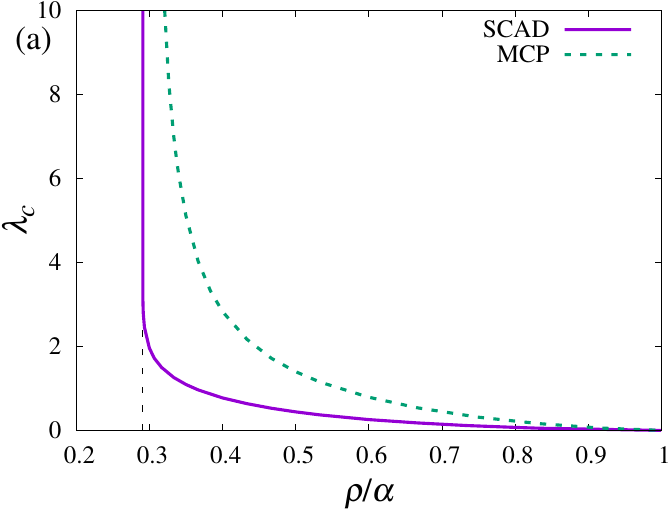}
\end{minipage}
\begin{minipage}{0.495\hsize}
\centering
\includegraphics[width=3in]{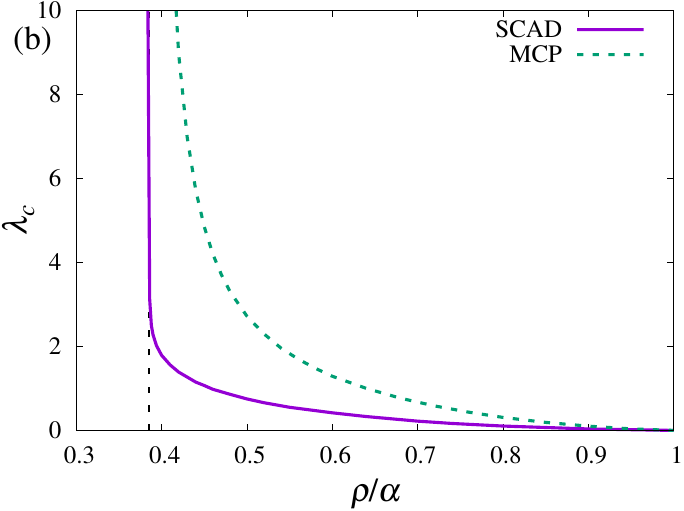}
\end{minipage}
\caption{$\rho\slash\alpha$-dependence of $\lambda_c$ for (a) $\alpha=0.3$
and (b) $\alpha=0.5$. 
In the parameter region on the left side to the vertical lines,
$\ell_1$ minimization reconstructs the original signals.}
\label{fig:lambda_c}
\end{figure}

We denote as $a_c(\lambda)$ the value of $a$ under which 
the signals can be reconstructed for each $\lambda$.
The reconstruction limit $a_c(\lambda)$ on the $\lambda-a$ plane is shown in 
\Fref{fig:phase_lambda_vs_a}
for (a) SCAD and (b) MCP, respectively, at $\alpha=0.5$ for $\rho=0.3$ and $\rho=0.4$.
The horizontal dashed lines represent $a_{\min}$, which is equal to 1 when the success solution is stable,
and the signals can be reconstructed in the parameter region $a_{\min}< a\leq a_c(\lambda)$.
The dashed vertical lines represent $\lambda_c$, defined as the maximum value of $\lambda$
that gives $a_c(\lambda)> a_{\min}$. Hence, 
the signal cannot be reconstructed at $\lambda\geq \lambda_c$.
For the reconstruction of dense signals,
small nonconvexity parameters $\lambda$ and $a$ are required,
and $a_c(\lambda)$ and $\lambda_c$ for MCP are always greater than that for SCAD.
The dependence of $\lambda_c$ on $\rho\slash\alpha$ for SCAD and MCP are compared 
in \Fref{fig:lambda_c} for (a) $\alpha=0.3$ and (b) $\alpha=0.5$.
The vertical lines represent the reconstruction limit of $\ell_1$ minimization, and
the values of $\lambda_c$ diverge as $\rho\slash\alpha$ approaches the $\ell_1$ reconstruction limit.
This divergence of $\lambda_c$ means that one can reconstruct the signals 
using any $\lambda\in(0,\infty)$ and $a\in(a_{\min},\infty)$
when the signals are sufficiently sparse 
to be reconstructed by $\ell_1$ minimization.
For any system parameters, 
the divergence of $\lambda_c$ in MCP is faster than that in SCAD,
which indicates that the range of
possible values of nonconvexity parameters
for reconstruction in MCP is wider than that in SCAD.
In this sense, MCP is superior to SCAD.

\begin{figure}
\begin{minipage}{0.495\hsize}
    \centering
    \includegraphics[width=3in]{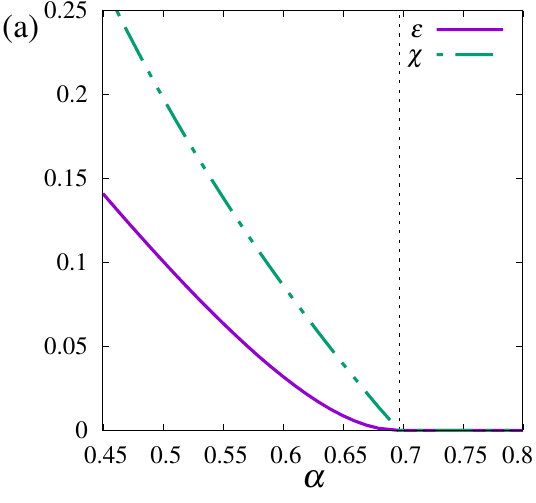}
\end{minipage}
\begin{minipage}{0.495\hsize}
\centering
    \includegraphics[width=3in]{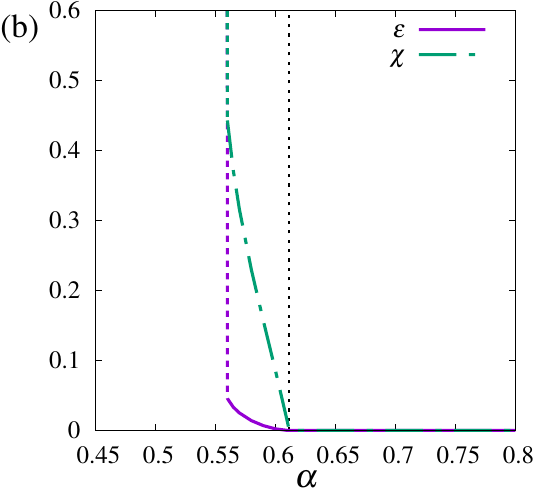}
\end{minipage}
    \caption{$\alpha$-dependence of $\varepsilon$ and $\chi$ in 
    RS solution of SCAD
    at $\rho=0.35$
    for (a) $\lambda=1,~a=10$ and 
    (b) $\lambda=0.3,~a=5$.
    {The vertical dotted lines indicate $\alpha_c$,
    and the vertical dashed lines in (b) indicate the disappearance of the
    finite $\varepsilon$ and $\chi$.}}
    \label{fig:SCAD_no_success}
\end{figure}
\begin{figure}
\begin{minipage}{0.495\hsize}
    \centering
    \includegraphics[width=3in]{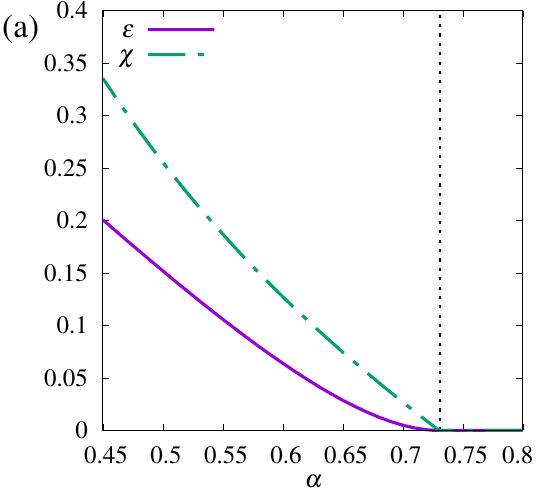}
\end{minipage}
\begin{minipage}{0.495\hsize}
\centering
    \includegraphics[width=3in]{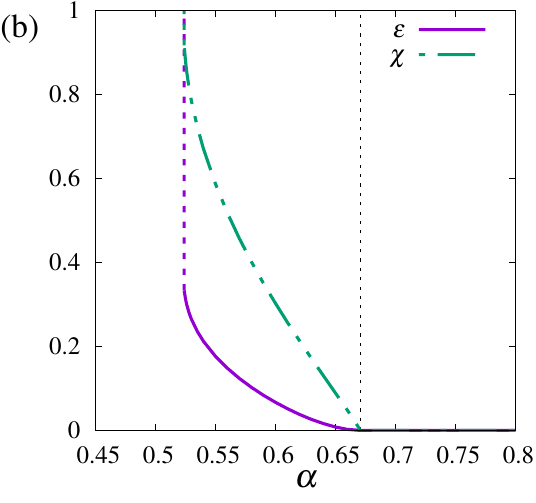}
\end{minipage}
    \caption{$\alpha$-dependence of $\varepsilon$ and $\chi$ in 
    RS solution of MCP
    at $\rho=0.4$
    for (a) $\lambda=1,~a=10$ and (b) $\lambda=0.5,~a=5$.
    {The vertical dotted lines indicate $\alpha_c$,
    and the vertical dashed lines in (b) indicate the disappearance of the
    finite $\varepsilon$ and $\chi$.}}
    \label{fig:MCP_no_success}
\end{figure}

\subsection{{Comment on the RS-failure solution}}
\label{sec:RS_failure}

We mention the existence of the solution of RS saddle point equation
at $\alpha<\alpha_c$
for subsequent discussions.
One can find a solution with $\varepsilon>0$ ($Q<\rho\sigma_x^2$ and $m<\rho\sigma_x^2$) and $\chi>0$ within $\Omega^\dag(a)$
at $\alpha<\alpha_c$,
which violates the AT condition.
We term this solution as {\it RS-unstable failure solution}.
Figs. \ref{fig:SCAD_no_success} and \ref{fig:MCP_no_success}
show $\alpha$-dependence of $\varepsilon$ and $\chi$ for SCAD and MCP,
respectively,
where the vertical dashed lines represent $\alpha_c$.
The RS-unstable failure solution
is smoothly connected to the success solution that appears at $\alpha\geq \alpha_c$,
{and do not coexist with the success solution.}
The RS-unstable failure solution
does not contribute to the equilibrium property of the system,
but this solution 
is still useful to consider the behavior of the algorithm
as shown in the following sections.

\subsection{{Comment on the diverging ``solution"}}

{
As shown in Figs. \ref{fig:SCAD_no_success} (b) and \ref{fig:MCP_no_success} (b),
when $\lambda$ is sufficiently small,
the values of $\varepsilon$ and $\chi$
tend to diverge at sufficiently small $\alpha$,
and the solution with finite $\varepsilon$ and $\chi$ disappears.
In fact, \eref{eq:chi_SCAD} indicates that 
the solution $\chi\to\infty$ is stable for both SCAD and MCP
when
\begin{eqnarray}
\alpha<(1-\rho)\mathrm{erfc}(\theta_\infty^-)+\rho\mathrm{erfc}(\theta_\infty^+)
\label{eq:diverging_stability}
\end{eqnarray}
holds, where $\theta_\infty^+=a\lambda\sqrt{\alpha\slash\{2(\alpha+\varepsilon)\}}$
and $\theta_\infty^-=a\lambda\sqrt{\alpha\slash(2\varepsilon)}$,
although the solution is out of the physical region $\Omega^\dag(a)$.
Considering the limit $\chi\to\infty$,
the set of the saddle point equations for SCAD and MCP is reduced to the same 
one equation for
the MSE $\varepsilon$ as
\begin{eqnarray}
\nonumber
\varepsilon&=\frac{\varepsilon(1-\rho)}{\alpha}\left\{\frac{2\theta_\infty^-}{\sqrt{\pi}}e^{-(\theta_\infty^-)^2}+\mathrm{erfc}(\theta_\infty^-)\right\}
+\frac{\rho(\varepsilon+\alpha)}{\alpha}\left\{\frac{2\theta_\infty^+}{\sqrt{\pi}}e^{-(\theta_\infty^+)^2}+\mathrm{erfc}(\theta_\infty^+)\right\}\\
&-2\rho\sigma_x^2\mathrm{erfc}(\theta_\infty^+)+\rho\sigma_x^2.
\label{eq:varepsilon_SCAD}
\end{eqnarray}
The solutions of \eref{eq:varepsilon_SCAD} and \eref{eq:varepsilon_MCP}
can be finite or infinite depending on $\alpha$, $\lambda$, $\rho$ and $a$,
and the infinite $\varepsilon$ is always stable if it exists when $\alpha<1$.
The diverging solutions do not contribute to the thermodynamic behavior,
because they are not in the region $\Omega^\dag(a)$,
but they affect the algorithmic behavior of SCAD or MCP minimization.
}

\section{Approximate message passing with nonconvexity control}
\label{sec:AMP}

For numerical computation of the estimate
under a given measurement matrix,
AMP is a feasible algorithm with
low computational cost.
As discussed below, the typical trajectory and fixed point of AMP
can be connected to the analysis based on replica method,
hence we can understand algorithmic behavior by comparing with the analysis.
The detailed derivation has been given by 
the previous studies \cite{Sakata-Xu,Krzakala2012,GAMP}, and 
here we briefly introduce the algorithm.
In AMP, the estimates under a general separable sparse penalty
is recursively updated as 
\begin{equation}
    \hat{x}_i^{(t+1)}=\arg\min_x\left\{\frac{\T{Q}_i^{(t)}}{2}x^2-h_i^{(t)}x+J(x_i;\lambda,a)\right\},
    \label{eq:single-body_AMP}
\end{equation}
where $\hat{x}_i^{(t)}$ denotes the estimate at time step $t$, and 
\begin{eqnarray}
\T{Q}_i^{(t)}&=\frac{1}{\hat{V}^{(t)}},\\
h_i^{(t)}&=\hat{x}_i^{(t-1)}\T{Q}_i^{(t)}+\sum_{\mu=1}^MA_{\mu i}R_\mu^{(t)},\\
\hat{V}^{(t)}&=\frac{1}{M}\sum_{i=1}^N\hat{v}_i^{(t-1)},\\
\hat{v}_i^{(t)}&=\frac{\partial \hat{x}_i^{(t)}}{\partial h_i^{(t)}},\\
R_\mu^{(t)}&=\frac{y_\mu-\sum_iA_{\mu i}\hat{x}_i^{(t-1)}}{\hat{V}^{(t)}}.
\end{eqnarray}
The solution of \eref{eq:single-body_AMP}
corresponds to the minimizer of \eref{eq:one_body},
with the replacement of $\T{Q}$ and $\sigma z$ with 
$\T{Q}_i^{(t)}$ and $h_i^{(t)}$, respectively.

\begin{figure}
\begin{minipage}{0.495\hsize}
    \centering
    \includegraphics[width=3in]
    {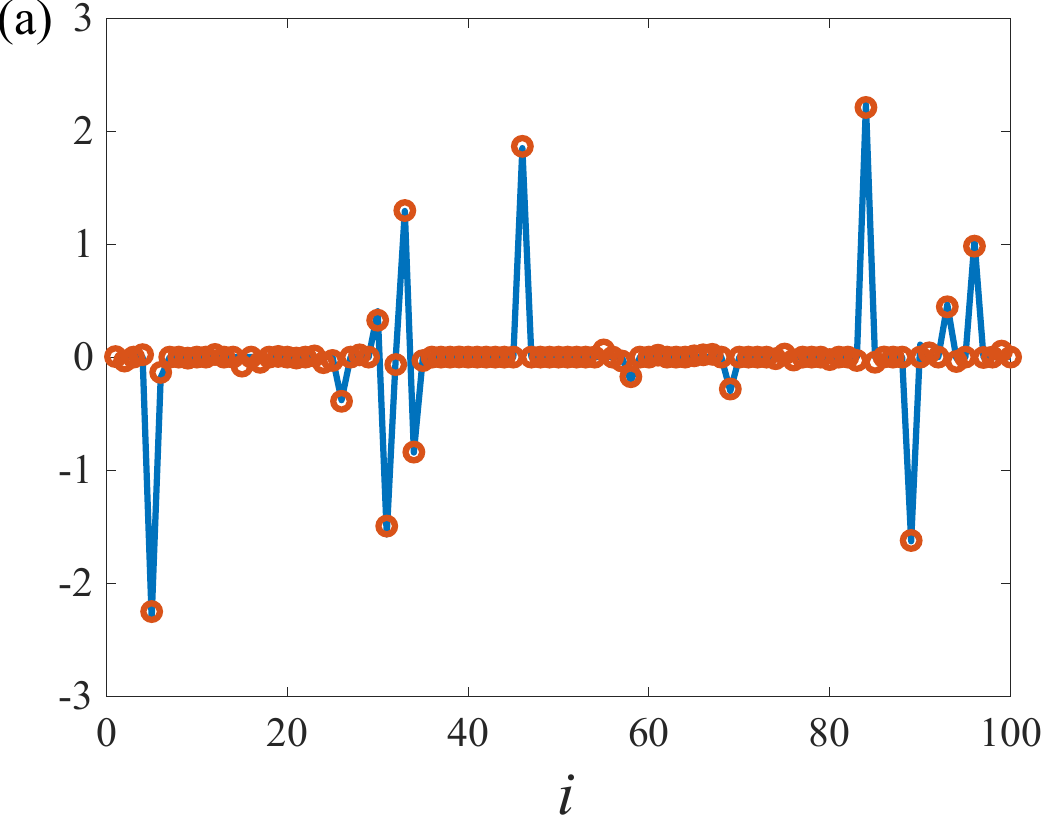}
    \end{minipage}
    \begin{minipage}{0.495\hsize}
    \centering
    \includegraphics[width=3in]
    {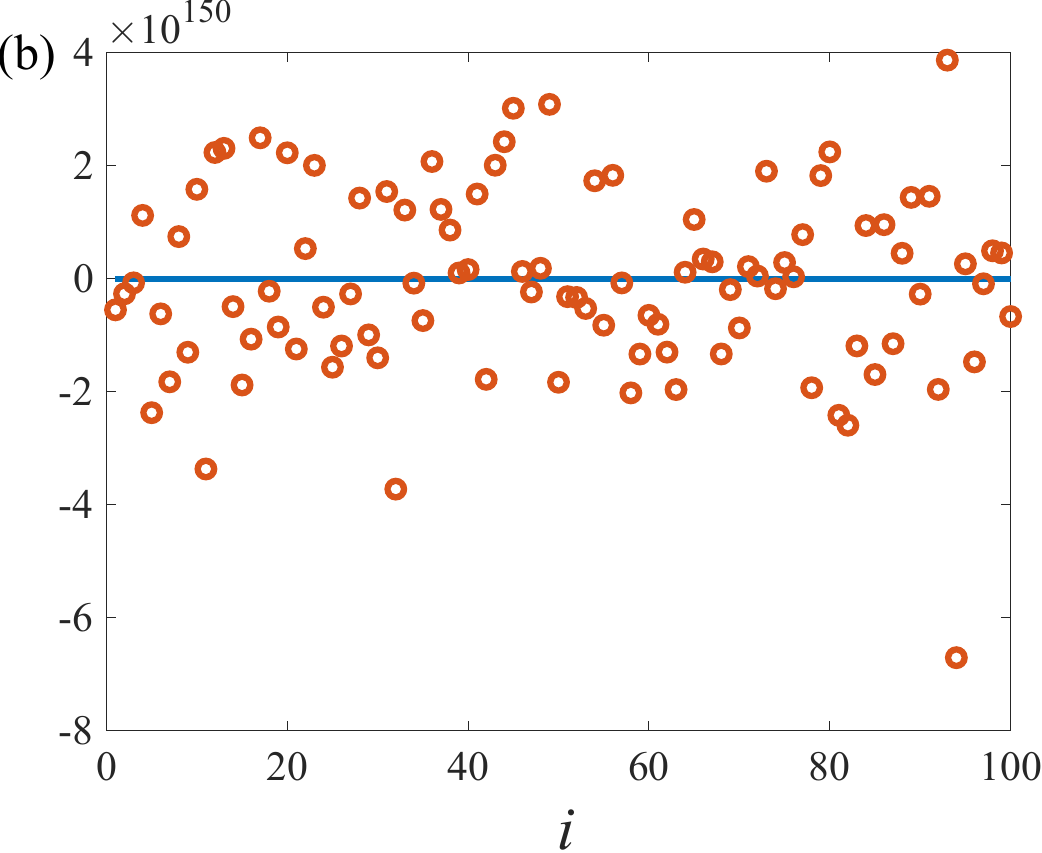}
    \end{minipage}
    \caption{
    True signal $x_i^0$ (solid line) and reconstructed signal $\hat{x}_i$ 
    (circles) at $N=100$ and $M=50$ 
    for (a) $\rho=0.25$ with SCAD at $\lambda=1$ and $a=3$,
    and (b) $\rho=0.28$ with SCAD at $\lambda=0.1$ and $a=3$.}
    \label{fig:no_anneal_X}
\end{figure}

The local stability of AMP corresponds to the AT instability condition \cite{Sakata-Xu}.
Hence, it is expected that AMP reconstructs the original signal
in the theoretically derived parameter region $\alpha> \alpha_c(\rho)$
at sufficiently large $N$,
because the current problem does not exhibit any first order transitions
or spinodal transitions
under fixed nonconvexity parameters,
in contrast to the Bayes-optimal setting \cite{Krzakala2012} or the 
Monte Carlo sampling case \cite{Obuchi_MC}.
\Fref{fig:no_anneal_X} shows examples of reconstructed signals of $N=400$ and $\alpha=0.5$
after 1000 steps update of AMP 
under SCAD (a) at $\lambda=1$ and $a=3$ for $\rho=0.25$
and (b) at $\lambda=0.1$ and $a=3$ for $\rho=0.3$,
where the original and reconstructed signals are 
represented by solid lines and circles, respectively.
In these parameter regions, perfect reconstruction is theoretically supported,
but as shown in \Fref{fig:no_anneal_X} (b), 
the naive update of AMP does not achieve the perfect reconstruction
for small values of nonconvexity parameters.
The discrepancy between the replica analysis and AMP appears, in particular, when 
the signal is dense. 
The tendency 
is common in both SCAD and AMP,
hence we explain their characteristic behavior 
using SCAD as a representative.

As mentioned above,
any other stable solutions do not exist
when the success solution is locally stable,
hence the
discrepancy between the replica analysis and AMP
cannot be understood by the spinodal transition as with the Bayes-optimal method.
For understanding the difficulty 
in achieving the perfect reconstruction of the signal by AMP at a small nonconvexity parameter region,
we use state evolution (SE) \cite{Mezard2009}.
The typical trajectory of AMP 
is characterized by two parameters
$V^{(t)}\equiv E_{\bm{x}^0,A}[\hat{V}^{(t)}]$ and $\varepsilon^{(t)}\equiv E_{\bm{x}^0,A}[\hat{\varepsilon}^{(t)}]$,
where $\hat{\varepsilon}^{(t)}\equiv \sum_{i=1}^N(\hat{x}_i^{(t)}-x_i^0)^2\slash N$ is the mean squared error at $t$-th iteration step.
In particular, when the components of $A$ are independently and identically distributed with
mean 0 and variance $1/N$, 
as for the Gaussian measurement matrix,
the time evolution of $V^{(t)}$ and $\varepsilon^{(t)}$
is described by SE equations \cite{Sakata-Xu,Krzakala2012}
\begin{eqnarray}
V^{(t+1)}&=\int dx^0 P_0(x^0)\int Dz \Sigma(\alpha^{-1}V^{(t)},x^0+z\sqrt{\alpha^{-1}\varepsilon^{(t)}}),\\
\varepsilon^{(t+1)}&=\int dx^0 P_0(x^0)\int Dz \left[\hat{x}(\alpha^{-1}V^{(t)},x^0+z\sqrt{\alpha^{-1}\varepsilon^{(t)}})-x^0\right]^2,
\end{eqnarray}
where $\hat{x}(s,w)=\Sigma
_{\mathrm{p}}(s,w)
{\cal M}_{\mathrm{p}}(s,w)$ for
$\mathrm{p}\in\{\mathrm{SCAD},\mathrm{MCP}\}$.
SE is equivalent to the RS saddle point equation,
and the fixed point denoted by $V^*$ and $\varepsilon^*$ correspond to 
the RS saddle point as $V^*=\chi$ and $\varepsilon^*=Q-2m+\rho\sigma_x^2$,
respectively.
Hence, the success solution is described as $V^*=\varepsilon^*=0$ in the SE.
{ 
As mentioned in Sec. \ref{sec:RS_failure},
the failure solution appears in some parameter regions, 
but it always involves the RS instability and never coexists with the success solution.}
Note that the flow of the SE describes the typical trajectory of the
AMP with respect to $A$ and $\bm{x}^0$.
Hence, it does not necessarily describe a trajectory
under a fixed realization of $A$ and $\bm{x}^0$.
However, it is expected that the trajectories converge to 
the flow of SE for a sufficiently large system size.
Hence, SE flow supports an understanding of a trajectory of AMP 
under a fixed set of $A$ and $\bm{x}^0$ \cite{Bayati-Montanari}.

\begin{figure}
\begin{minipage}{0.495\hsize}
\includegraphics[width=3in]{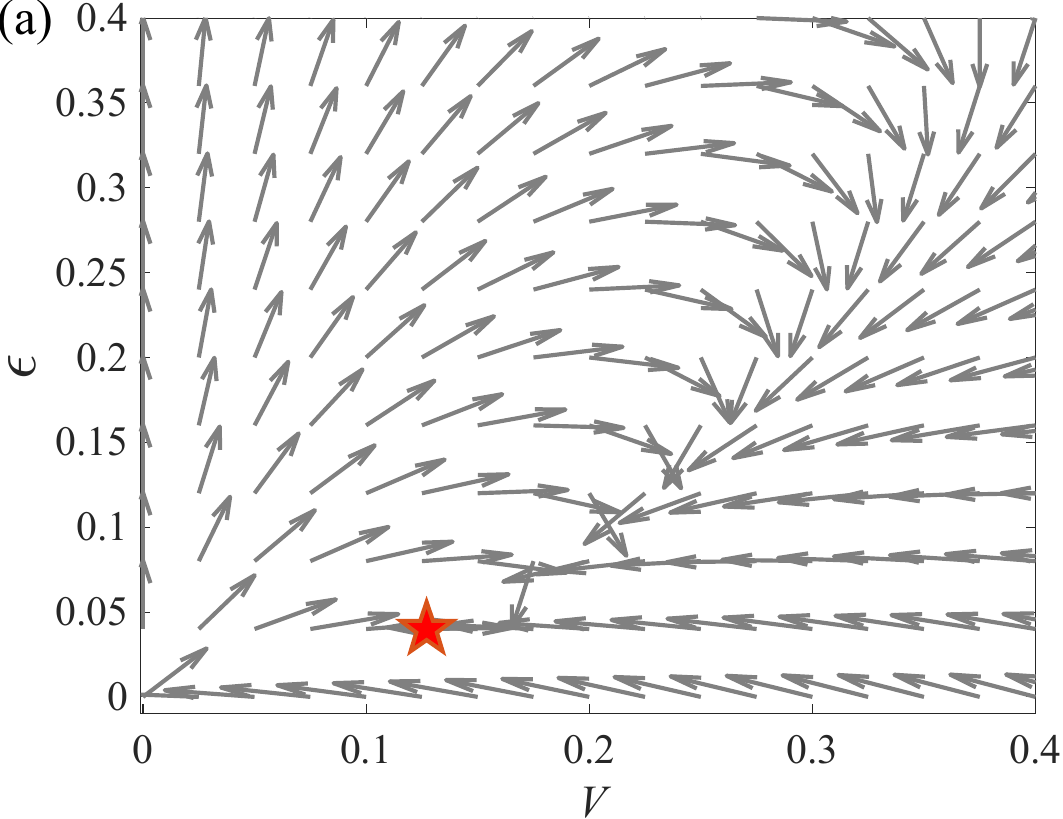}
\end{minipage}
\begin{minipage}{0.495\hsize}
\includegraphics[width=3in]{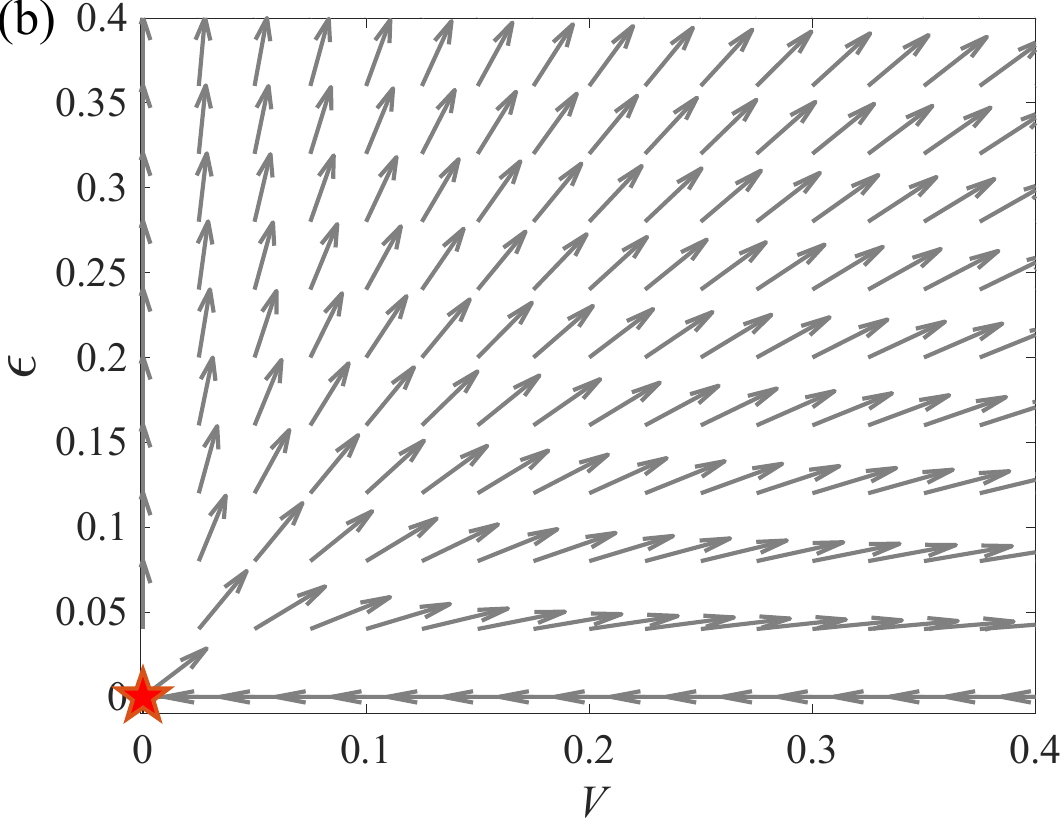}
\end{minipage}
\caption{Flow of state evolution at $\alpha=0.5$ and $\rho=0.28$
for SCAD (a) $\lambda=1$ and $a=3$ and
(b) $\lambda=0.1$ and  $a=3$.
The fixed points are depicted by stars.}
\label{fig:SE_flow}
\end{figure}

\Fref{fig:SE_flow} shows the flow of SE at 
$\alpha=0.5$ and $\rho=0.28$
for SCAD at (a) $\lambda=1$ and $a=3$ and 
(b) $\lambda=0.1$ and $a=3$.
The arrows assigned to the coordinate $(\hat{V},\hat{\varepsilon})$
is the 
normalized vector of 
$(V^{(t+1)}-V^{(t)},\varepsilon^{(t+1)}-\varepsilon^{(t)})$
with $V^{(t)}=\hat{V}$ and $\varepsilon^{(t)}=\hat{\varepsilon}$,
which indicate the direction of SE's flow,
and the stars depict the fixed points of 
SE.
As shown in \Fref{fig:SE_flow} (a),
SE has a fixed point with finite $\varepsilon$ and $V$
for large $\lambda$, which corresponds to the 
RS-unstable failure solution,
but as $\lambda$ decreases,
most of the SE flow leads to a divergence of $V$ and $\varepsilon$ as shown in
\Fref{fig:SE_flow}(b);
however, there is still a region close to the $V$-axis where the flows are directed to $V=\varepsilon=0$.
This region shrinks to $V$-axis as the nonconvexity parameter decreases.
Flows of SE in MCP is almost the same as SCAD shown in \Fref{fig:SE_flow}.
In case of $\ell_1$ minimization, one can check that
SE reaches to the success solution from 
any initial condition of $(V,\varepsilon)$, namely 
the volume of the basin of attraction (BOA) diverges to infinity
at $\alpha>\alpha_c(\rho)$ or $\rho<\rho_c(\alpha)$.
Therefore, this 
shrinking basin {and the diverging flow are significant properties} of the 
minimization problems of SCAD and MCP.

\begin{figure}
\begin{minipage}{0.495\hsize}
\includegraphics[width=3in]{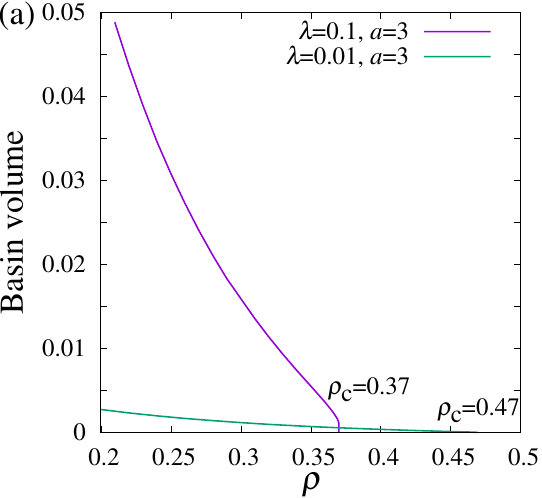}
\end{minipage}
\begin{minipage}{0.495\hsize}
\includegraphics[width=3in]{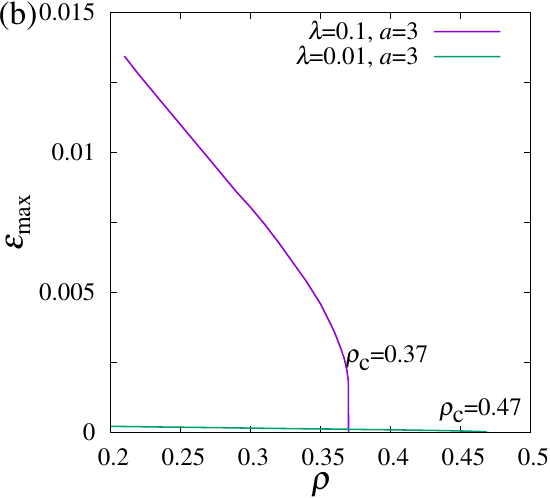}
\end{minipage}
\caption{(a) Volume of the BOA to $V=\varepsilon=0$ at $\alpha=0.5$.
(b) Maximum value of $\varepsilon$, denoted by $\varepsilon_{\max}$, 
on the boundary of the BOA.}
\label{fig:basin}
\end{figure}

We quantify the volume of the BOA under SCAD minimization
and show its dependency on $\rho$ for $\alpha=0.5$ as 
\Fref{fig:basin} (a).
The BOA to $V=\varepsilon=0$ 
is zero at $\rho=\rho_c$, and gradually increases from zero
as $\rho$ decreases from $\rho_c$.
When the nonconvexity parameter $\lambda$ is small,
the basin volume tends to be small for any $\rho$ region.
\Fref{fig:basin}(b) shows the possible maximum value of $\varepsilon$,
denoted by $\varepsilon_{\max}$,
as an initial condition to converge to $V=\varepsilon=0$.
Namely, $\varepsilon_{\max}$ is the maximum value of $\varepsilon$
on the boundary of the BOA.
It means that to achieve perfect reconstruction
at sufficiently large $\rho$ with small nonconvexity parameter,
we need to set initial condition as $\varepsilon\sim O(10^{-2})$ for $\lambda=0.1$ and $a=3$,
and as $\varepsilon\sim O(10^{-5})$ for $\lambda=0.01$ and $a=3$.
Such initial conditions with small $\varepsilon$ is not realistic, and
the possibility that AMP 
attains the success solution $V=\varepsilon=0$ from randomly chosen initial conditions
is exceedingly small.

The shrinking BOA is the origin of the 
difficulty of AMP for small nonconvexity parameters.
To resolve this problem,
we recall that 
the RS-unstable failure solution appears for $\alpha<\alpha_c$
for large nonconvexity parameters, as discussed in Figs. \ref{fig:SCAD_no_success}
and \ref{fig:MCP_no_success}.
The emergence of the RS-unstable failure solution implies
that the SE has a locally stable fixed point
from the correspondence between the saddle point of RS free energy and the SE,
as shown in \Fref{fig:SE_flow}(a).
Therefore, AMP for large nonconvexity parameter 
does not converge to a fixed point,
but its trajectory is confined into a subshell characterized by 
finite $\varepsilon$ and $V$.
We utilize this nondivergence property of AMP in $\alpha<\alpha_c$ 
at sufficiently large nonconvexity parameters
for the perfect reconstruction of dense signals 
at small nonconvexity parameters.
The procedure introduced here, based on
the above consideration, is termed as {\it nonconvexity control},
where we decrease the value of nonconvexity parameters
in updating AMP.

\begin{figure}
\begin{minipage}{0.495\hsize}
    \centering
    \includegraphics[width=3in]
    {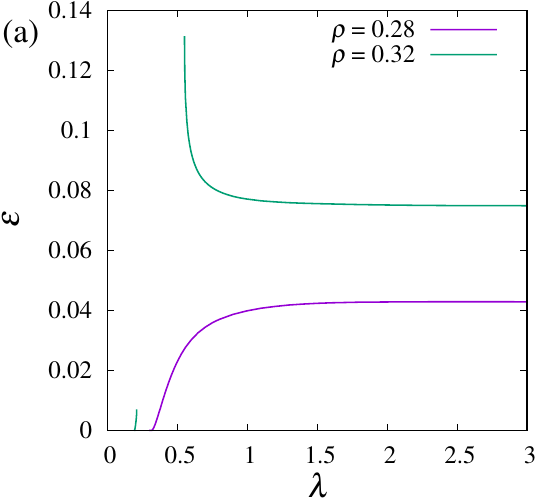}
    \end{minipage}
    \begin{minipage}{0.495\hsize}
    \centering
    \includegraphics[width=3in]
    {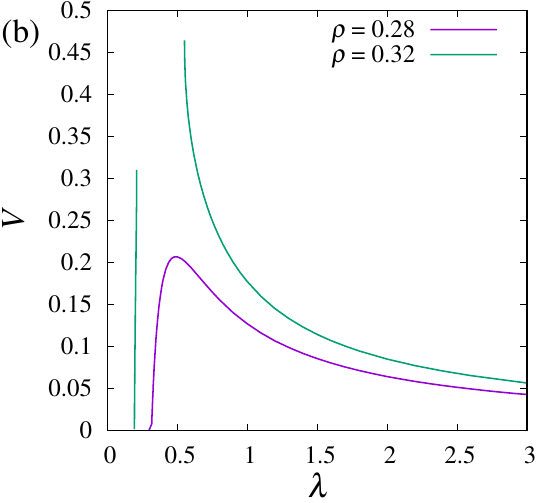}
    \end{minipage}
    \caption{{$\lambda$-dependence of (a) $\varepsilon$ and (b) $V$ at 
    $\alpha=0.5$ and $a=3$ for $\rho=0.28$ and $\rho=0.32$.}}
    \label{fig:V_and_E}
\end{figure}

\begin{figure}
\centering
\includegraphics[width=4in]{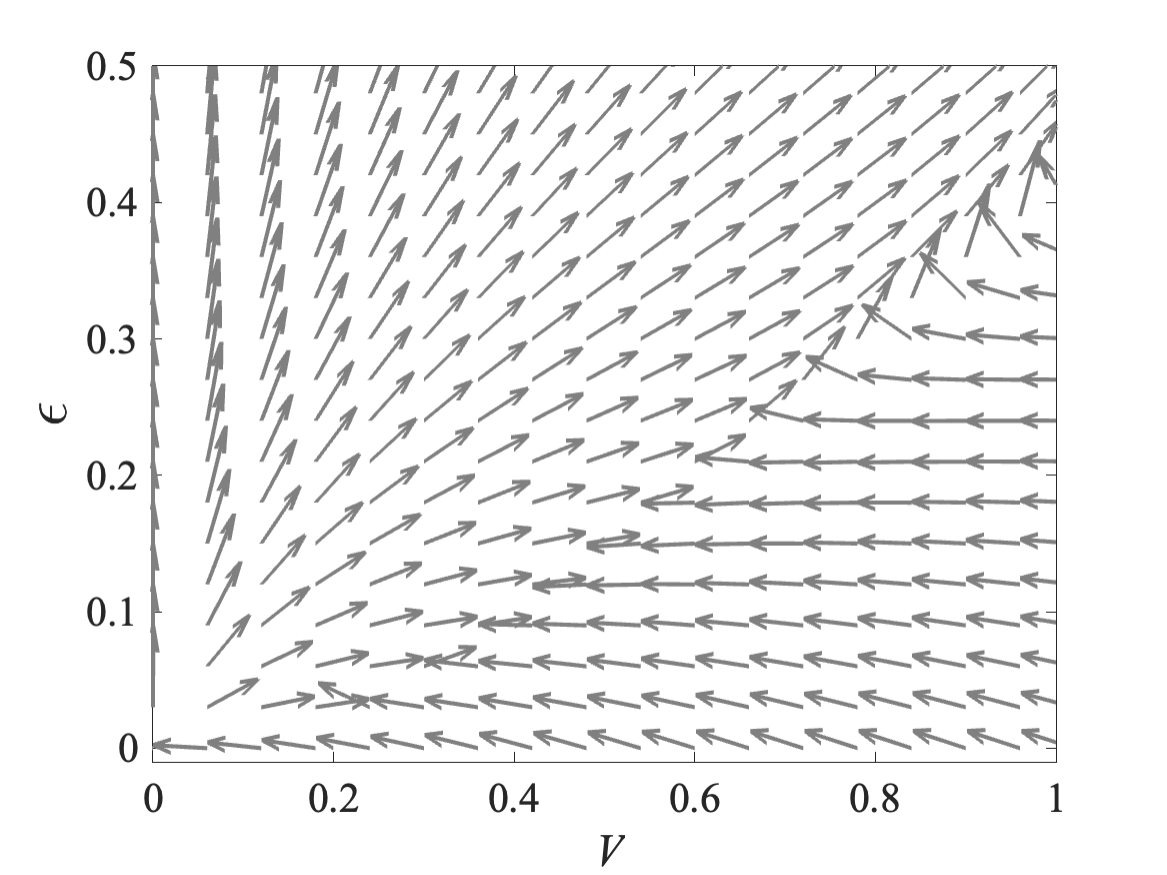}
\caption{{SE flow at $\alpha=0.5$, $\rho=0.32$, $\lambda=0.4$ and $a=3$.
There is no fixed point and the flow diverges.}}
\label{fig:SE_divergence}
\end{figure}

{
Here, we consider the control of the parameter $\lambda$
under a fixed value of $a$.
\Fref{fig:V_and_E} shows
$\lambda$-dependence of $\varepsilon$ and $V$ at 
$\alpha=0.5$ and $a=3$ for $\rho=0.28$ and $\rho=0.32$,
and explains how the nonconvexity control proposed here works 
or fails for the perfect reconstruction.
We treat the set of macroscopic fixed points as a
sequence generated by shifting the value of $\lambda$.
Note that the sequence mainly consists of RS-unstable failure solutions.
At $\rho=0.28$, the sequences of $\varepsilon$ and $V$ 
are connected to zero by decreasing $\lambda$, hence
one can potentially attain perfect reconstruction
by starting from large $\lambda$ and decreasing $\lambda$.
However, at $\rho=0.32$,
$\varepsilon$ and $V$ tend to diverge when $\lambda\in(0.2117,0.5530)$
as shown in \Fref{fig:V_and_E}.
In this region, the finite RS-unstable fixed points are disappeared,
and the SE flow goes to the diverging state, although 
the state is not allowed as a solution.
\Fref{fig:SE_divergence} is an example of SE flow 
going towards the diverging state in the absence of the solution with finite $\varepsilon$ and $V$,
which is observed at $\alpha=0.5, \rho=0.32,\lambda=0.4$ and $a=3$.
The discontinuity in the sequence of the macroscopic parameters 
and the SE flow to the diverging state
obstructs the nonconvexity control.
}

\begin{figure}
\begin{minipage}{0.495\hsize}
    \centering
    \includegraphics[width=3in]
    {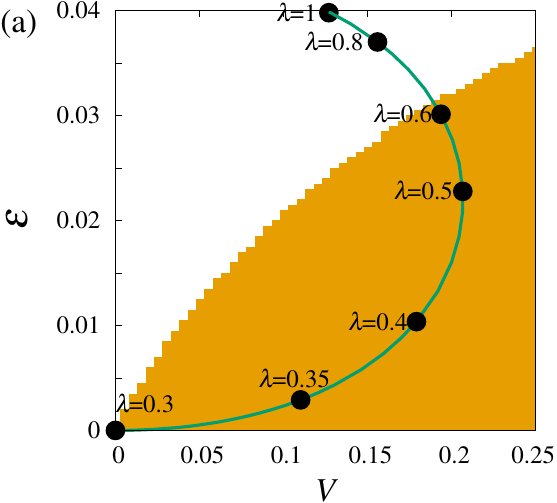}
    \end{minipage}
    \begin{minipage}{0.495\hsize}
    \centering
    \includegraphics[width=3in]
    {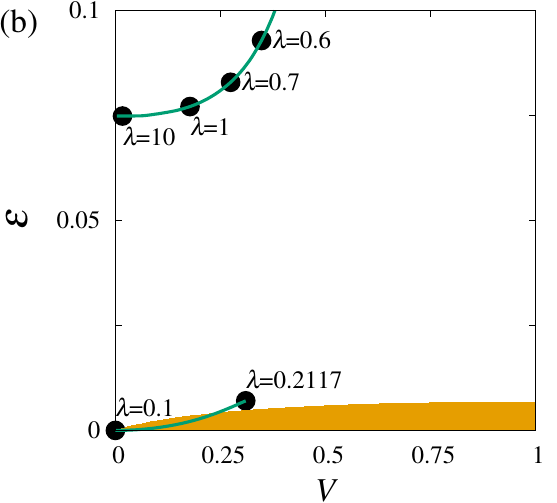}
    \end{minipage}
    \caption{Sequence of macroscopic
    fixed points of SCAD corresponding to continuous decrease in
    $\lambda$ at $a=3$ (solid line) and the
    BOA (shaded region) for a sufficiently small $\lambda$ at $\alpha=0.5$.
    Dotts denote representative fixed points at the $\lambda$ assigned next to the dotts.
    (a) Macroscopic fixed points and the BOA at $\lambda=0.3$ is shown for $\rho=0.28$.
    (b) Macroscopic fixed points and the BOA at $\lambda=0.1$ is shown for $\rho=0.32$.}
    \label{fig:basin_and_FP}
\end{figure}

{In \Fref{fig:basin_and_FP},
we compare the sequence of the macroscopic fixed point
with the BOA to the origin
at $\alpha=0.5$ and $a=3$ for (a) $\rho=0.28$ and (b) $\rho=0.32$.
The solid lines of \Fref{fig:basin_and_FP}
are drawn by continuously shifting $\lambda$,
and are equivalent to \Fref{fig:V_and_E}.
Dots on the lines represent examples of the fixed points at each $\lambda$.}
In \Fref{fig:basin_and_FP} (a),
the shaded region denotes the BOA to 
$V=\varepsilon=0$ at $\lambda=0.3$ and $a=3$, 
where the perfect reconstruction is 
theoretically supported.
To attain perfect reconstruction at this parameter region,
we need to prepare the initial condition with $\varepsilon\sim O(10^{-2})$,
which is not realistic.
However, by decreasing the nonconvexity parameter $\lambda$
from larger values such as $\lambda=1$, 
the fixed point, which corresponds to RS-unstable failure solution,
comes into the BOA to the perfect reconstruction
at $\lambda=0.3$ as shown in \Fref{fig:basin_and_FP} (a).
{
In \Fref{fig:basin_and_FP} (b),
BOA to the perfect reconstruction at 
$\lambda=0.1$ and $a=3$ is depicted as the shaded region
at $\alpha=0.5$ and $\rho=0.32$.
For larger $\rho$, the sequence of the fixed point shows discontinuity
as shown in \Fref{fig:basin_and_FP}(b),
which is caused by the diverging property of the macroscopic quantities
as shown in Figs. \ref{fig:V_and_E} and \ref{fig:SE_divergence}.}
It is expected that the sequence of the fixed point below $\lambda=0.2117$
can provide a clue to attain the perfect reconstruction,
but corresponding BOA is already shrunk.

\begin{figure}
\begin{minipage}{0.495\hsize}
\centering
\includegraphics[width=3in]{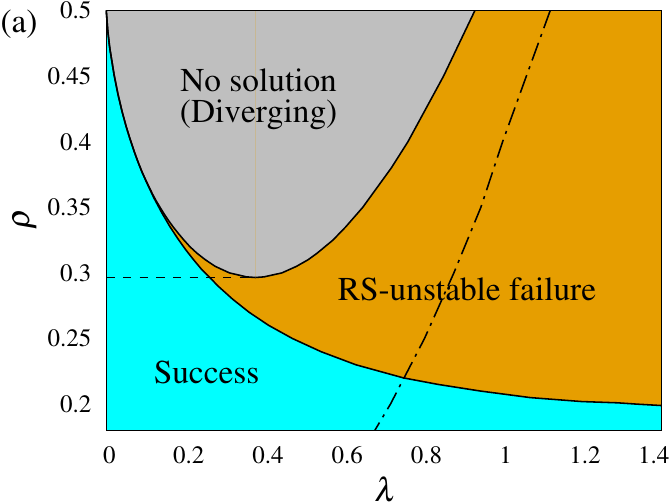}
\end{minipage}
\begin{minipage}{0.495\hsize}
\includegraphics[width=3in]{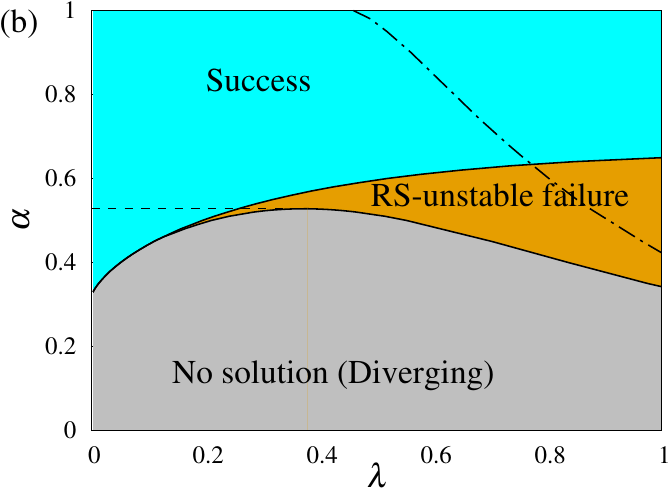}
\end{minipage}
\caption{{
Phase diagram on (a) $\rho-\lambda$ plane at $\alpha=0.5$ and $a=3$
and (b) $\alpha-\lambda$ plane at $\rho=0.32$ and $a=3$.
The horizontal dashed lines denote the nonconvexity control limit 
(a) $\rho_{\mathrm{NCC}}(\alpha=0.5)$ and 
(b) $\alpha_{\mathrm{NCC}}(\rho=0.32)$.
On the left of the dotted-dashed line,
the diverging state is stable and the SE flow tends to diverge.}}
\label{fig:phase_rho_vs_lambda}
\end{figure}

{
Based on the abovementioned observations,
we define {\it nonconvexity control limit} (NCC limit) $\rho_{\mathrm{NCC}}(\alpha)$
as the largest value of $\rho$ 
under given $\alpha$ at which 
the sequence of the fixed points reach 
$V=\varepsilon=0$ as $\lambda$ decreases 
without facing to the discontinuity due to the divergence of the 
macroscopic parameters; 
$\alpha_{\rm NCC}(\rho)$ is defined as well. 
\Fref{fig:phase_rho_vs_lambda}(a) and (b)
show the phase diagram on $\rho-\lambda$
plane at $\alpha=0.5$ and $a=3$, and that on $\alpha-\lambda$ plane 
at $\rho=0.32$ and $a=3$.
The NCC limit is denoted by the horizontal dashed lines,
and the stability of the diverging state 
\eref{eq:diverging_stability} is satisfied on the left side of dotted-dashed lines.
At $\rho<\rho_{\mathrm{NCC}}$ or $\alpha>\alpha_{\mathrm{NCC}}(\rho)$,
the sequence of the fixed point is connected to $V=\varepsilon=0$
by decreasing $\lambda$ as shown in \Fref{fig:basin_and_FP} (a).
Examples of the SE's flow below the NCC limit
in the RS-unstable failure and the success region
are shown in \Fref{fig:SE_flow} (a) and (b), respectively.
At $\rho>\rho_{\mathrm{NCC}}$ or $\alpha<\alpha_{\mathrm{NCC}}$,
``No solution'' region, in which SE does not have any fixed points
in $\Omega^\dag(a)$ and diverges, appears between 
``Success" region and ``RS-unstable failure" region,
and the nonconvexity control fails due to the ``No solution" region.
An example of the SE flow in the ``No solution'' region
is \Fref{fig:SE_divergence}.
As $\lambda$ increases, the RS-unstable failure phase and 
the success phase are connected to each other for any $\rho$.
This property is the same as the $\ell_1$ minimization.
}

\begin{figure}
\begin{minipage}{0.495\hsize}
    \centering
    \includegraphics[width=3.2in]
    {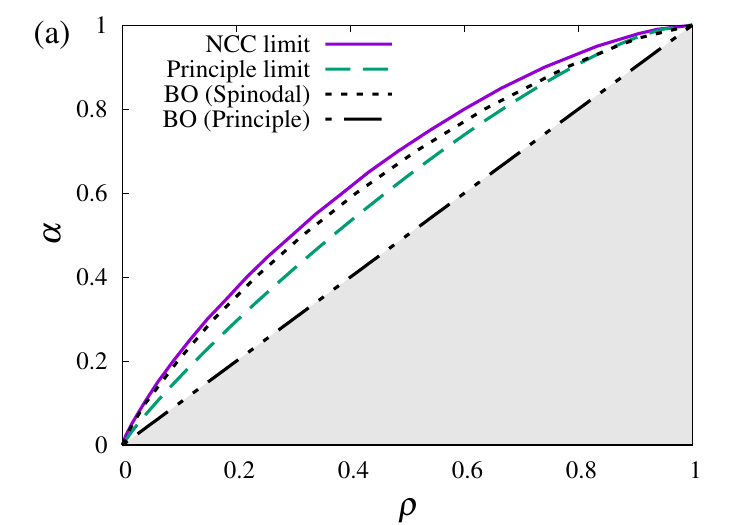}
    \end{minipage}
    \begin{minipage}{0.495\hsize}
    \includegraphics[width=3.2in]
    {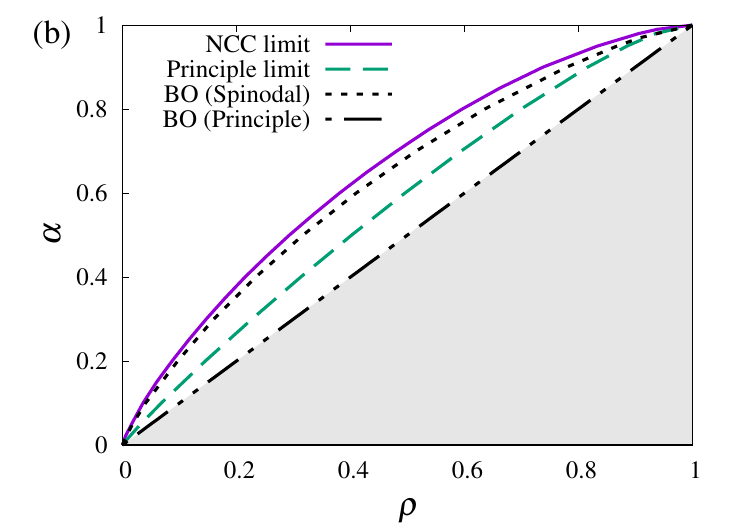}
    \end{minipage}
    \caption{
    Reconstruction limit by nonconvexity control (NCC limit) and 
    the phase transition boundary (Principle limit) at $\lambda=0.1$ and $a=3$
    for (a) SCAD and (b) MCP. 
    For comparison, the algorithmic limit of Bayes optimal method (BO, Spinodal) and
    the principle limit of Bayes optimal method (BO, Principle) are shown.}
    \label{fig:NCC_phase}
\end{figure}

{
\Fref{fig:NCC_phase} shows the 
$\alpha$-dependence of the NCC limit $\rho_{\mathrm{NCC}}(\alpha)$
for (a) SCAD and (b) MCP.
The dependency of $\rho_{\mathrm{NCC}}$ 
on $a$ is weak; the changes in $a$ 
induce the changes in the $\rho$ value smaller than $O(10^{-3})$,
and here the value is maximized with respect to $a$.
For comparison, the phase transition boundary $\alpha_c(\rho)$ for $\lambda=0.1$
and $a=3$
is shown as principle limit in the sense that 
the stability of the success solution is guaranteed.
According to the principle limit, 
it is expected that MCP can achieve perfect reconstruction under denser signals
than SCAD, 
but its NCC limit inferior to SCAD in the order $O(10^{-2})$ in terms of $\rho$.
This observation implies that there remains a room for improvement 
in designing nonconvex penalties to overcome the basin shrinkage
sustaining global stability of $V=\varepsilon=0$ in dense region.}

\begin{figure}
\begin{minipage}{0.495\hsize}
\centering
\includegraphics[width=3in]{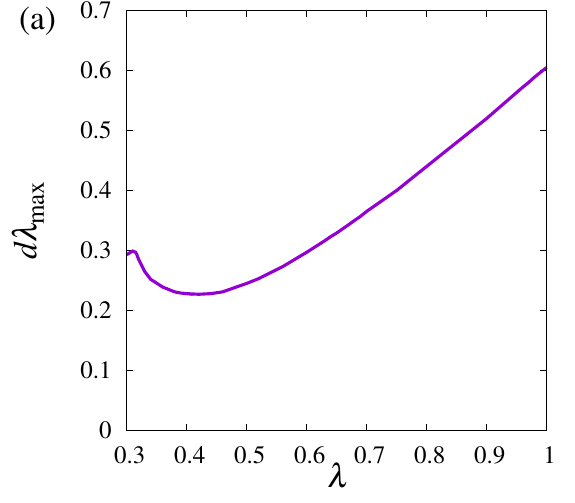}
\end{minipage}
\begin{minipage}{0.495\hsize}
\centering
\includegraphics[width=3in]{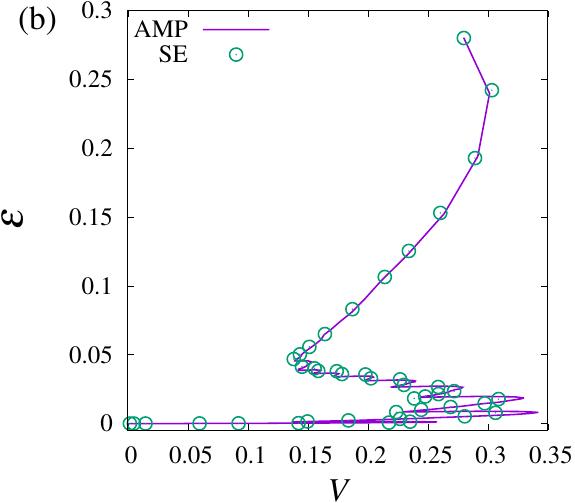}
\end{minipage}
\caption{(a) The value of $d\lambda_{\max}$ below which the
nonconvexity control is effective at $\alpha=0.5$, $\rho=0.28$ and $a=3$.
(b) Trajectory on $V-\varepsilon$ plane with convexity control of SCAD.
The trajectories of AMP for one realization of $\bm{X}^0$ and $A$
at $N=10^5$ (solid line) and 
SE (circles) are shown.
The initial condition is set to be $V=E=\rho$,
and $\lambda=1$, $a=3$.
The value of $\lambda$ is decreased by $d\lambda=0.1$ 
after the convergence of $\hat{V}^{(t)}$ and $D^{(t)}$ at each $\lambda$.}
\label{fig:SCAD_annealing}
\end{figure}

The problem in practice
is the protocol of the nonconvexity control.
We consider here an `equilibrium' approach;
we spend sufficient time steps at each $\lambda$
for the convergence to the macroscopic fixed point, which
corresponds to the RS-unstable failure solution, 
and after that we decrease $\lambda$ by $d\lambda$.
Hence, we need to set the sufficient time for `equilibration' and $d\lambda$ appropriately.
However,  we cannot assess $\hat{\varepsilon}^{(t)}$
in AMP's trajectory since its calculation requires the unknown true signal $\bm{x}^0$.
Instead, we observe $\hat{V}^{(t)}$ and $\hat{D}^{(t)}\equiv\frac{1}{N}\sum_{i=1}^N(\hat{x}_i^{(t+1)}-\hat{x}_i^{(t)})^2$
as criteria of the convergence,
and we decrease $\lambda$ by $d\lambda$ after the saturation of the 
$\hat{V}^{(t)}$ and $\hat{D}^{(t)}$ around certain values.
Next, in determining $d\lambda$,
the macroscopic fixed point at $\lambda$ is required to be
in the BOA of the macroscopic fixed point at $\lambda-d\lambda$
for the effective nonconvexity control.
We denote the maximum value of $d\lambda$ as $d\lambda_{\max}$
over which the abovementioned condition is violated,
and choose a value $d\lambda$ smaller than $d\lambda_{\max}$
for nonconvexity control.
The value of $d\lambda_{\max}$ assessed by SE
at $\alpha=0.5$ and $\rho=0.28$
is shown as the solid line in 
\Fref{fig:SCAD_annealing}(a),
where $d\lambda_{\max}$ is obtained 
by observing the SE flow at $\lambda-d\lambda$ starting 
from the fixed point at $\lambda$ for various $d\lambda$.
Here, we note that at this parameter region,
the perfect reconstruction is possible at $\lambda<0.3$,
hence $d\lambda$ for $\lambda<0.3$ is trivially $d\lambda\sim\lambda$.

The value of $d\lambda$ 
shown in \Fref{fig:SCAD_annealing}(a) is for the 
typical realization of $A$ and $\bm{x}^0$, and 
the possible value of $d\lambda$ might fluctuate
depending on $A$ and $\bm{x}^0$.
To be on the safe side,
we set $d\lambda=0.1$ for any $\lambda$
in applying the nonconvexity control to AMP under a given $A$ and $\bm{x}^0$.
\Fref{fig:SCAD_annealing}(b)
shows the actual trajectory of AMP (solid line) for one realization of
$A$ and $\bm{x}^0$ at $N=10^5$,
and corresponding SE (circles)
at $\alpha=0.5$, $\rho=0.28$ under nonconvexity control.
The initial condition of AMP is set to be $\bm{x}=\bm{0}$ and $\bm{v}=\rho\bm{1}_N$,
where $\bm{1}_N$ is the $N$-dimensional vector whose all components are 1,
and hence that of SE is $V=E=\rho$.
We start with $\lambda=1$ at $a=3$, and decrease $\lambda$ by 
$d\lambda=0.1$ after the convergence of $\hat{V}^{(t)}$ and $\hat{D}^{(t)}$
at each $\lambda$,
until $\lambda$ becomes to be 0.3.
The behavior of AMP is well described by SE.
Comparing to the flow of SE (\Fref{fig:SE_flow} (b)),
the trajectory of AMP with nonconvexity control 
approaches the BOA connected to the success solution at $\lambda=0.1$
that is almost on the $V$ axis. 

\section{Summary and discussion}
\label{sec:summary}

We have analytically derived the perfect reconstruction limit of the sparse signal in
compressed sensing by the minimization of
nonconvex sparse penalties, SCAD and MCP. 
In particular, when the nonconvexity parameters are small,
SCAD and MCP minimization reconstruct dense signals
that are beyond the   
$\ell_1$ reconstruction limit.
This analytical result also appears 
to imply that SCAD and MCP minimization overcomes 
the algorithmic limit of the Bayes optimal method, 
but the numerical experiments using AMP have shown that this is actually not the case. 
The gap between the analytical and numerical results has been understood by observing
the flow of SE, revealing the failure of AMP comes from
the shrinking BOA and the divergent behavior of AMP in some parameter region. 
We have found that SCAD and MCP minimization show a novel failure scenario of the algorithm
different from the Bayes-optimal setting where 
the algorithmic limit is characterized by the emergence of local minima. 
To mitigate the abovementioned gap and determine the algorithmic limit of AMP, 
we have proposed the protocol of the nonconvexity control, 
leading to a largely improved performance.

Originally, SCAD and MCP were designed to satisfy the continuity and the oracle property,
which is the simultaneous appearance of
the asymptotic normality and consistency,
at a certain limit with respect to the nonconvexity parameters \cite{SCAD,MCP}.
However, such property is not sufficient for practical usage.
The design of the nonconvex penalties that does not 
lead to a shrinkage of basin is 
another possibility for nonconvex compressed sensing.
From the relationship between 
the sparse prior in Bayesian approach and sparse penalty in frequentist approach,
it is implied that SCAD and MCP can be related to the
Bernoulli-Gaussian prior in Bayesian terminology with 
large variance \cite{Bayes_L0}.
Unified understanding of the sparsity over Bayesian and frequentist approach
will be helpful for designing such desirable sparse penalties.
Further, the 
application of the nonconvexity control to the general matrix
beyond that consists of i.i.d. entries should be discussed for
practical usage.
The rotationally invariant matrix is one of the candidates
to examine the effectiveness of the nonconvexity control 
for general matrix \cite{VAMP,Krzakala2020,Takahashi}.


\ack

The authors would like to thank Yoshiyuki Kabashima, 
Satoshi Takabe, Mirai Tanaka, and Yingying Xu 
for their helpful discussions and comments.
This work is partially supported by JSPS KAKENHI No. 19K20363 (AS), 
and Nos. 19H01812, 18K11463 and 17H00764 (TO), 
Japan Science and Technology Agency (JST) 
PRESTO Grant No. JPMJPR19M2 (AS), and
a Grant for Basic Science Research Projects from the Sumitomo Foundation (TO).

\section*{References}


\providecommand{\newblock}{}

\end{document}